\documentclass{article}

\usepackage[final, nonatbib]{neurips_2021}

\usepackage{listings}
\usepackage{enumitem}
\usepackage[utf8]{inputenc} 
\usepackage[T1]{fontenc}    
\usepackage{url}            
\usepackage{booktabs}       
\usepackage{amsfonts}       
\usepackage{nicefrac}       
\usepackage{amssymb}
\usepackage{microtype}      
\usepackage{xcolor}         

\usepackage{amsmath,bm}
\usepackage[utf8]{inputenc} 
\usepackage[T1]{fontenc}    
\usepackage{hyperref}       
\usepackage{url}            
\usepackage{booktabs}       
\usepackage{amsfonts}       
\usepackage{nicefrac}       
\usepackage{microtype}      
\usepackage{url,hyphenat}
\usepackage{caption,subfigure,graphicx,epstopdf,multirow,multicol,booktabs,verbatim,wrapfig,bm}
\usepackage{makecell}
\usepackage{amsthm,amssymb,amsfonts,amsmath}
\usepackage{algorithmic}

\usepackage{bbding} 
\usepackage{threeparttable}
\usepackage{multirow}
\usepackage{multicol}
\usepackage{subfigure}
\usepackage[vlined,boxed,commentsnumbered,linesnumbered,ruled]{algorithm2e}
\newtheorem{definition}{Definition}
\newtheorem{assumption}{Assumption}
\newtheorem{theorem}{Theorem}

\newtheorem{corollary}{Corollary}
\newtheorem{proposition}{Proposition}
\newtheorem{remark}{Remark}

\title{From Canonical Correlation Analysis to Self-supervised Graph Neural Networks}
\author{
  Hengrui Zhang$^{1}$\thanks{This work was done during the author's internship at AWS Shanghai AI Lab.} , Qitian Wu$^{2}$, Junchi Yan$^{2}$, David Wipf$^{3}$, Philip S. Yu$^{1}$\thanks{Corresponding author.} 
  \\
  $^1$ Department of Computer Science, University of Illinois at Chicago \\
  $^2$ Department of Computer Science and Engineering, Shanghai Jiao Tong University\\
   $^3$AWS Shanghai AI Lab \\
  \texttt{{hzhan55}@uic.edu, \{echo740, yanjunchi\}@sjtu.edu.cn}\\ \texttt{daviwipf@amazon.com, psyu@uic.edu} \\

}
\begin{document}

\maketitle

\begin{abstract}
We introduce a conceptually simple yet effective model for self-supervised representation learning with graph data. It follows the previous methods that generate two views of an input graph through data augmentation. However, unlike contrastive methods that focus on instance-level discrimination, we optimize an innovative feature-level objective inspired by classical Canonical Correlation Analysis. Compared with other works, our approach requires none of the parameterized mutual information estimator, additional projector, asymmetric structures, and most importantly, negative samples which can be costly. We show that the new objective essentially 1) aims at discarding augmentation-variant information by learning invariant representations, and 2) can prevent degenerated solutions by decorrelating features in different dimensions. Our theoretical analysis further provides an understanding for the new objective which can be equivalently seen as an instantiation of the Information Bottleneck Principle under the self-supervised setting. Despite its simplicity, our method performs competitively on seven public graph datasets. The code is available at: \url{https://github.com/hengruizhang98/CCA-SSG}.
\end{abstract}

\section{Introduction}
Self-supervised learning (SSL) has been a promising paradigm for learning useful representations without costly labels~\cite{bert, cpc, simclr}. In general, it learns representations via a proxy objective between inputs and self-defined signals, among which contrastive methods~\cite{cpc, cmc, moco, simclr, byol} have achieved impressive performance on learning image representations by maximizing the mutual information of two views (or augmentations) of the same input. Such methods can be interpreted as a discrimination of a joint distribution (positive pairs) from the product of two marginal ones (negative pairs) \cite{hypersphere}.

Inspired by the success of contrastive learning in vision~\cite{dim, cpc, cmc, simclr,moco, byol, simsiam}, similar methods have been adapted to learning graph neural networks~\cite{dgi, mvgrl, gcc, grace, grace-ad}. Although these models have achieved impressive performance, they require complex designs and architectures. For example, DGI~\cite{dgi} and MVGRL~\cite{mvgrl} rely on a parameterized mutual information estimator to discriminate positive node-graph pairs from negative ones; GRACE~\cite{grace} and GCA~\cite{grace-ad} harness an additional MLP-projector to guarantee sufficient capacity. Moreover, negative pairs sampled or constructed from data often play an indispensable role in providing effective contrastive signals and have a large impact on performance. Selecting proper negative samples is often nontrivial for graph-structured data, not to mention the extra storage cost for prohibitively large graphs. BGRL~\cite{bgrl} is a recent endeavor on targeting a negative-sample-free approach for GNN learning through asymmetric architectures~\cite{byol, simsiam}. However, it requires additional components, e.g., an exponential moving average (EMA) and Stop-Gradient, to empirically avoid degenerated solutions, leading to a more intricate architecture. 

Deviating from the large body of previous works on contrastive learning, in this paper we take a new perspective to address SSL on graphs. We introduce Canonical Correlation Analysis inspired Self-Supervised Learning on Graphs (CCA-SSG), a simple yet effective approach that opens the way to a new SSL objective and frees the model from intricate designs. It follows the common practice of prior arts, generating two views of an input graph through random augmentation and acquiring node representations through a shared GNN encoder. Differently, we propose to harness a \emph{non-contrastive} and \emph{non-discriminative} feature-level objective, which is inspired by the well-studied Canonical Correlation Analysis (CCA) methods~\cite{cca, cca-computation, multi-cca, kernel-cca, deep-cca, soft-decorrelation}. More specifically, the new objective aims at maximizing the correlation between two augmented views of the same input and meanwhile decorrelating different (feature) dimensions of a single view's representation. We show that the objective 1) essentially pursuits discarding augmentation-variant information and preserving augmentation-invariant information, and 2) can prevent dimensional collapse~\cite{decorrelation} (i.e., different dimensions capture the same information) in nature. Furthermore, our theoretical analysis sheds more lights that under mild assumptions, our model is an instantiation of Information Bottleneck Principle~\cite{IB-1,IB-2, IB-3} under SSL settings~\cite{barlow, Multi-IB, mini-1}.

To sum up, as shown in Table~\ref{tbl-comparison}, our new objective induces a simple and light model without reliance on negative pairs~\cite{dgi, mvgrl, grace, grace-ad}, a parameterized mutual information estimator~\cite{dgi, mvgrl}, an additional projector or predictor~\cite{grace, grace-ad, bgrl} or asymmetric architectures~\cite{bgrl, mvgrl}. We provide a thorough evaluation for the model on seven node classification benchmarks. The empirical results demonstrate that despite its simplicity, CCA-SSG can achieve very competitive performance in general and even superior test accuracy in five datasets. It is worth noting that our approach is agnostic to the input data format, which means that it can potentially be applied to other scenarios beyond graph-structured data (such as vision, language, etc.). We leave such a technical extension for future works.
    
\textbf{Our contributions are as follows:}
\begin{itemize}[topsep=0in,leftmargin=0em,wide=0em]

\item[\textbf{1)}]  We introduce a non-contrastive and non-discriminative objective for self-supervised learning, which is inspired by Canonical Correlation Analysis methods. It does not rely on negative samples, and can naturally remove the complicated components. Based on it we propose CCA-SSG, a simple yet effective framework for learning node representations without supervision (see Section~\ref{sec:method}).

\item[\textbf{2)}] We theoretically prove that the proposed objective aims at keeping augmentation-invariant information while discarding augmentation-variant one, and possesses an inherent relationship to an embodiment of Information Bottleneck Principle under self-supervised settings (see Section~\ref{sec:insights}).

\item[\textbf{3)}] Experimental results show that without complex designs, our method outperforms state-of-the-art self-supervised methods MVGRL~\cite{mvgrl} and GCA~\cite{grace-ad} on 5 out of 7 benchmarks. We also provide thorough ablation studies on the effectiveness of the key components of CCA-SSG (see Section~\ref{sec:experiments}). 
\end{itemize}

\begin{table}[tb!]
	\centering
	\caption{Technical comparison of self-supervised node representation learning methods. We provide a conceptual comparison with more self-supervised methods in Appendix~\ref{append:compare}. \textit{Target} denotes the comparison pair, N/G/F denotes node/graph/feature respectively.
	\textit{MI-Estimator}: parameterized mutual information estimator. \textit{Proj/Pred}: additional (MLP) projector or predictor. \textit{Asymmetric}: asymmetric architectures such as EMA and Stop-Gradient, or two separate encoders for two branches. \textit{Neg examples}: requiring negative examples to prevent trivial solutions. \textit{Space} denotes space requirement for storing all the pairs. Our method is simple without any listed component and memory-efficient.
}
	\label{tbl-comparison}
	\small
	\begin{threeparttable}
    {{
    		\begin{tabular}{c|lcccccc}
			\toprule[0.8pt]
			& Methods & Target & MI-Estimator & Proj/Pred & Asymmetric & Neg examples & Space  \\
			\midrule[0.6pt]
           \multirow{6}{*}{\rotatebox{90}{Instance-level}} 
            & DGI~\cite{dgi}                & N-G  & \Checkmark & - & - & \Checkmark & $O(N)$   \\
            & MVGRL~\cite{mvgrl}  & N-G & \Checkmark & - & \Checkmark & \Checkmark & $O(N)$  \\
            & GRACE~\cite{grace}  & N-N & - & \Checkmark & - & \Checkmark & $O(N^2)$ \\
            & GCA~\cite{grace-ad}  & N-N & - & \Checkmark & - & \Checkmark & $O(N^2)$ \\
            & BGRL~\cite{bgrl}     & N-N & - & \Checkmark &  \Checkmark & - & $O(N)$ \\
            \midrule[0.5pt]
            & CCA-SSG (Ours)      & F-F &- & - & - & - & $O(D^2)$ \\
			\bottomrule[0.8pt]
		\end{tabular}
		}
		}
	\end{threeparttable}
\end{table}

\section{Related Works and Background}

\paragraph{Contrastive Learning on Graphs.}
Contrastive methods~\cite{cpc, cmc,dim,moco,simclr,byol} have been shown to be effective for unsupervised learning in vision, which have also been adapted to graphs. Inspired by the local-global mutual information maximization viewpoints~\cite{dim}, DGI~\cite{dgi} and InfoGraph~\cite{infograph} put forward unsupervised schemes for node and graph representation learning, respectively. MVGRL~\cite{mvgrl} generalizes CMC~\cite{cmc} to graph-structured data by introducing graph diffusion~\cite{diffusion} to create another view for a graph. GCC~\cite{gcc} adopts InfoNCE loss~\cite{cpc} and MoCo-based negative pool~\cite{moco} for large-scale GNN pretraining. GRACE~\cite{grace}, GCA~\cite{grace-ad} and GraphCL~\cite{graphcl} follow the spirit of SimCLR~\cite{simclr} and learn node/graph representations by directly treating other nodes/graphs as negative samples. BGRL~\cite{bgrl} targets a negative-sample-free model, inspired by BYOL~\cite{byol}, on node representation learning. But it still requires complex asymmetric architectures.
\paragraph{Feature-level Self-supervised Objectives.}
The above-mentioned methods all focus on instance-level contrastive learning. To address their drawbacks, some recent works have been turning to feature-level objectives. For example, Contrastive Clustering~\cite{contrast-cluster} regards different feature dimensions as different clusters, thus combining the cluster-level discrimination with instance-level discrimination. W-MSE~\cite{whiten} performs a differentiable whitening operation on learned embeddings, which implicitly scatters data points in embedding space. Barlow Twins~\cite{barlow} borrows the idea of redundancy reduction and adopts a soft decorrelation term that makes the cross-correlation matrix of two views' representations close to an identity matrix. By contrast, our method is based on the classical Canonical Correlation Analysis, working by correlating the representations of two views from data augmentation and meanwhile decorrelating different feature dimensions of each view's representation.

\paragraph{Canonical Correlation Analysis.} CCA is a classical multivariate analysis method, which is first introduced in~\cite{cca}. For two random variables $X \in \mathbb{R}^{m}$ and $Y \in \mathbb{R}^{n}$, their covariance matrix is $\Sigma_{XY}= Cov(X,Y)$. CCA aims at seeking two vectors $a \in \mathbb{R}^{m} $ and $b \in \mathbb{R}^{n} $ such that the correlation $ \rho = \text{corr} (a^{\top}X,b^{\top}Y) = \frac{a^{\top}\Sigma_{XY}b}{\sqrt{a^{\top}\Sigma_{XX}a} \sqrt{b^{\top}\Sigma_{YY}b} } $ is maximized. Formally, the objective is
\begin{equation}
    \max\limits_{a,b} a^{\top}\Sigma_{XY}b, \; \mbox{s.t.} \; a^{\top}\Sigma_{XX}a = b^{\top}\Sigma_{YY} b = 1.
\end{equation}
For multi-dimensional cases, CCA seeks two sets of vectors maximizing their correlation and subjected to the constraint that they are uncorrelated with each other~\cite{cca-computation}. Later studies apply CCA to multi-view learning with deep models~\cite{deep-cca, multi-cca, kernel-cca}, by replacing the linear transformation with neural networks. Concretely, assuming $X_1, X_2$ as two views of an input data, it optimizes
\begin{equation}
    \mathop{\max}\limits_{\theta_1, \theta_2} \text{Tr}\left(P_{\theta_1}^{\top}(X_1)P_{\theta_2}(X_2)\right) \; \mbox{s.t.} \; P_{\theta_1}^{\top}(X_1)P_{\theta_1}(X_1) =  P_{\theta_2}^{\top}(X_2)P_{\theta_2}(X_2) = I.
\end{equation}
where $P_{\theta_1}$ and $P_{\theta_2}$ are two feedforward neural networks and $I$ is an identity matrix. Despite its preciseness, such computation is really expensive~\cite{soft-decorrelation}. Fortunately, soft CCA~\cite{soft-decorrelation} removes the hard decorrelation constraint by adopting the following Lagrangian relaxation:
\begin{equation}
    \mathop{\min}\limits_{\theta_1, \theta_2} \mathcal{L}_{dist}\left(P_{\theta_1}(X_1), P_{\theta_2}(X_2)\right) + \lambda\left(\mathcal{L}_{SDL}(P_{\theta_1}(X_1)) + \mathcal{L}_{SDL}(P_{\theta_2}(X_2))\right),
\end{equation}
where $\mathcal{L}_{dist}$ measures correlation between two views' representations and $\mathcal{L}_{SDL}$ (called stochastic decorrelation loss) computes an $L_1$ distance between $P_{\theta_i}(X_i)$ and an identity matrix, for $i=1, 2$. 

\section{Approach}\label{sec:method}
\subsection{Model Framework}\label{sec:framework}
\begin{figure}[tb!]
    \centering
    \includegraphics[width = 1.0\linewidth]{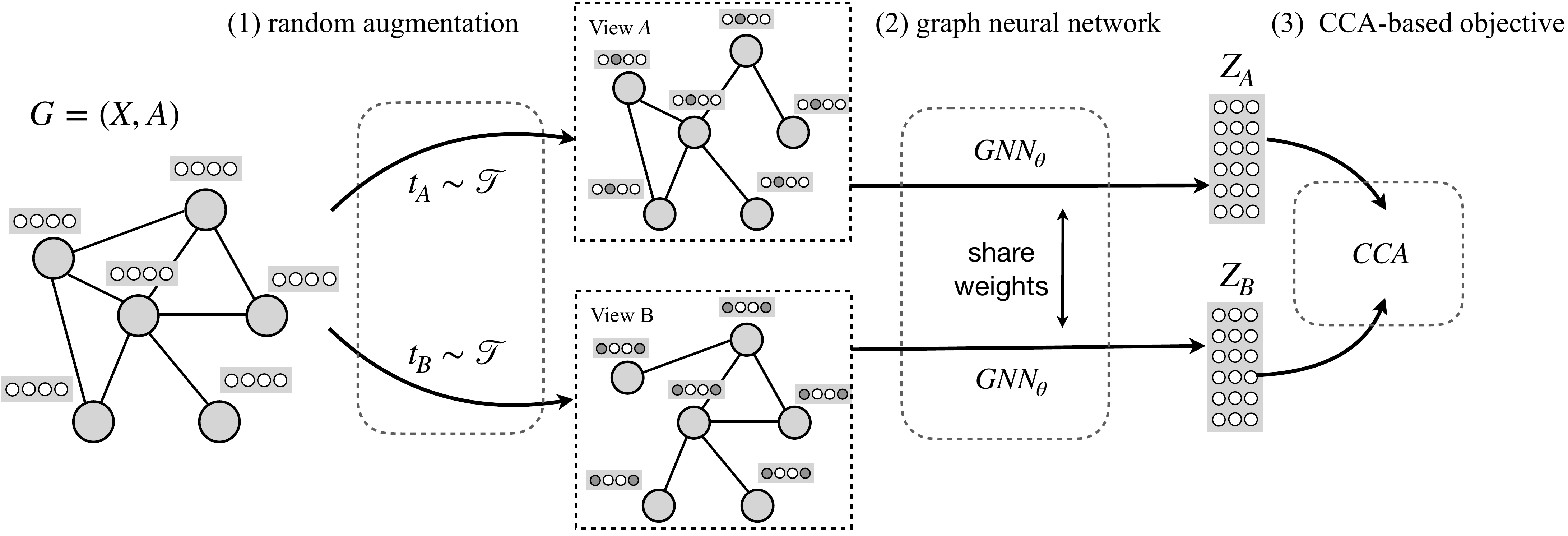}
    \caption{Illustration of the proposed model: given an input graph, we first generate two views through random augmentations: edge dropping and node feature masking. The two views are subsequently put into a shared GNN encoder to generate representations. The loss function is applied on the column-normalized embedding matrix of the two views. Note that this simple yet effective pipeline can also be conceptually applied for other data like vision and texts, which we leave for future works.}
    \label{fig:illustration}
    \vspace{-10pt}
\end{figure}

In this paper we focus on self-supervised node representation learning, where we consider a single graph $\mathbf{G} = (\mathbf{X}, \mathbf{A})$. $\mathbf{X}\in \mathbb{R}^{N\times F}$ and $\mathbf{A} \in \mathbb{R}^{N\times N}$ denote node features and adjacency matrix respectively. Here $N$ is the number of nodes within the graph and $F$ denotes feature dimension.

Our model simply consists of three parts: 1) a random graph augmentation generator $\mathcal{T}$. 2) a GNN-based graph encoder $f_{\theta}$ where $\theta$ denotes its parameters. 
3) a novel feature-level objective function based on Canonical Correlation Analysis. Fig.~\ref{fig:illustration} is an illustration of the proposed model.

 \begin{wrapfigure}{L}{0.53\textwidth}
    \begin{minipage}{0.53\textwidth}
      \begin{algorithm}[H]
        \caption{PyTorch-style code for CCA-SSG}
        \label{alg-torch}
        \definecolor{codeblue}{rgb}{0.25,0.5,0.5}
        \definecolor{codekw}{rgb}{0.85, 0.18, 0.50}
        \lstset{
          backgroundcolor=\color{white},
          basicstyle=\fontsize{7.5pt}{7.5pt}\ttfamily\selectfont,
          columns=fullflexible,
          breaklines=true,
          captionpos=b,
          commentstyle=\fontsize{7.5pt}{7.5pt}\color{codeblue},
          keywordstyle=\fontsize{7.5pt}{7.5pt}\color{codekw},
        }
\begin{lstlisting}[language=python]
# f: encoder network
# lambda: trade-off
# D: embedding dimension
# g: input graph
# feat: node features

# generate two views through random augmentation
g1, feat1 = augment(g, feat)
g2, feat2 = augment(g, feat)
z1 = f(g1, feat1)   # embedding of the 1st view
z2 = f(g2, feat2)   # embedding of the 2st view

# batch normalization
z1_norm = ((z1 - z1.mean(0)) / z1.std(0))/ sqrt(D)
z2_norm = ((z2 - z2.mean(0)) / z2.std(0))/ sqrt(D)

# covariance matrix of each view
c1 = torch.mm(z1_norm.T(), z1_norm)
c2 = torch.mm(z2_norm.T(), z2_norm)

iden = torch.eye(D)
loss_inv = (z1_norm - z2_norm).pow(2).sum()
loss_dec_1 = (c1 - iden).pow(2).sum()
loss_dec_2 = (c2 - iden).pow(2).sum()
loss_dec = loss_dec_1 + loss_dec_2
loss = loss_inv + lambda * loss_dec
\end{lstlisting}
\vspace{-7pt}
  \end{algorithm}
\end{minipage}
\vspace{-10pt}
\end{wrapfigure}
\textbf{Graph augmentations}. We consider the standard pipeline for random graph augmentation that has been commonly used in previous works 
\cite{grace, bgrl}. To be specific, we harness two ways for augmentation: \textbf{edge dropping} and \textbf{node feature masking}. Edge dropping randomly drops a fraction of edges from the original graph, while node feature masking randomly masks a fraction of features for all the nodes. In this way, $\mathcal T$ is composed of all the possible graph transformation operations and each $t\sim \mathcal T$ denotes a specific graph transformation for graph $G$. Note that we use commonly adopted augmentation methods to stay our focus on the design of objective function and conduct fair comparison with existing approaches. More complicated random augmentations~\cite{graphcl, grace-ad} can also be readily plugged into our model. Details for the used augmentation functions are in Appendix~\ref{append:exp}.

\textbf{Training}. In each training iteration, we first randomly sample two graph transformations $t_A$ and $t_B$ from $\mathcal{T}$, and then generate two views $\tilde{\mathbf{G}}_A = (\tilde{\mathbf{X}}_A, \tilde{\mathbf{A}}_A)$ and $\tilde{\mathbf{G}}_B = (\tilde{\mathbf{X}}_B, \tilde{\mathbf{A}}_B)$ according to the transformations. The two views are subsequently fed into a shared GNN encoder to generate the node embeddings of the two views: $\mathbf{Z}_A = f_{\theta}(\tilde{\mathbf{X}}_A, \tilde{\mathbf{A}}_A)$, $\mathbf{Z}_B = f_{\theta}(\tilde{\mathbf{X}}_B, \tilde{\mathbf{A}}_B)$, where $\mathbf{Z}_A, \mathbf{Z}_B \in \mathbb{R}^{N \times D}$ and $D$ denotes embedding dimension. We further normalize the node embeddings along instance dimension so that each feature dimension has a 0-mean and $1/\sqrt{N}$-standard deviation distribution: 
\begin{equation}
    \tilde{{\mathbf{Z}}} = \frac{\mathbf{Z} - \mu(\mathbf{Z})}{ \sigma(\mathbf{Z}) *\sqrt{N}}
\end{equation}
The normalized $\tilde{\mathbf{Z}}_A$, $\tilde{\mathbf{Z}}_B$ will be used to compute a feature-level objective in Section~\ref{sec:objective}. To help better understand the proposed framework, we provide the PyTorch-style pseudocode for training CCA-SSG in Algorithm~\ref{alg-torch}.

\textbf{Inference}. To generate node embeddings for downstream tasks, we put the original graph $\mathbf{G} = (\mathbf{X}, \mathbf{A})$ into the trained graph neural network $f_{\theta}$ and obtain node embeddings $\mathbf{Z} = f_{\theta}(\mathbf{X}, \mathbf{A})$.

\subsection{Learning Objective}\label{sec:objective}
Canonical Correlation Analysis has shown its great power in multi-view learning like instance recognition~\cite{soft-decorrelation}. However, it still remains unexplored to leverage CCA for self-supervised learning. Note that in SSL, one generates two sets of data from the same input through transformation or random data augmentation, which could be regraded as two views of the input data. This inspires us to introduce the following objective for self-supervised representation learning:
\begin{equation}\label{eqn:loss}
    \mathcal{L} = \underbrace{\left\| \tilde{\mathbf{Z}}_A -  \tilde{\mathbf{Z}}_B \right\|_F^2}_{\text{invariance term}}
     + \lambda \underbrace{\left( \left\|\tilde{\mathbf{Z}}_A^{\top}\tilde{\mathbf{Z}}_A - \mathbf{I}\right\|_F^2 + \left\|\tilde{\mathbf{Z}}_B^{\top}\tilde{\mathbf{Z}}_B - \mathbf{I}\right\|_F^2 \right)}_{\text{decorrelation term}}
\end{equation}
where $\lambda$ is a non-negative hyperparameter trading off two terms. Note that minimizing the invariance term is essentially maximizing the correlation between two views as their representations are already normalized. In SSL, as the two augmented views come randomly from the same distribution, we can adopt one encoder $f_{\theta}$ that is shared across two branches and seek for a regularization that encourages different feature dimensions to capture distinct semantics via the decorrelation term.

We next provide a variance-covariance perspective to  the new objective, following similar lines of reasoning in \cite{tian-byol, tian-dual}. Assume that input data come from a distribution $\bm{x} \sim p(\bm{x})$ and $\bm{s}$ is a view of $\bm{x}$ through random augmentation $\bm{s} \sim p_{aug}(\cdot|\bm{x})$. Denote $\bm{z}_{\bm{s}}$ as the representation of $\bm{s}$, then minimizing the invariance term, by expectation, is to minimize the variance of the normalized representation $\tilde{\bm{z}}_{\bm{s}}$, conditioned on $\bm{x}$. Also, minimizing the decorrelation term is to push the off-diagonal elements of the covariance matrix (given by two $\tilde{\bm{z}}_{\bm{s}}$'s) close to $0$. Formally, we have
\begin{equation}\label{eqn:loss-inv}
    \mathcal{L}_{inv} = \left\| \tilde{\mathbf{Z}}_A - \tilde{\mathbf{Z}}_B \right\|_F^2 =  \sum\limits_{i=1}^{N}\sum\limits_{k=1}^D(\tilde{z}^A_{i,j}- \tilde{z}^B_{i,j})^2 \cong {\mathbb{E}}_{\bm{x}}\left[\sum\limits_{k=1}^D \mathbb{V}_{\bm{s}|\bm{x}}[\bm{\tilde{z}}_{s, k}] \right] * 2N,
\end{equation}
\begin{equation}\label{eqn:loss-decor}
    \mathcal{L}_{dec} = \left\|\tilde{\mathbf{Z}}_S^{\top}\tilde{\mathbf{Z}}_S  - \mathbf{I}\right\|_F^2 = \left\| \text{Cov}_{\bm{s}}[\tilde{\bm{z}}] - I \right\|_F^2 \cong \sum\limits_{i\neq j} \left(\rho^{\bm{z}_s}_{i,j}\right)^2, \; \mbox{for}\; \tilde{\mathbf{Z}}_S\in \{\tilde{\mathbf{Z}}_A, \tilde{\mathbf{Z}}_B\},
\end{equation}
where $\rho$ is the Pearson correlation coefficient.
\subsection{Advantages over Contrastive Methods}
In this subsection we provide a systematic comparison with previous self-supervised methods for node representation learning, including DGI~\cite{dgi}, MVGRL~\cite{mvgrl}, GRACE~\cite{grace}, GCA~\cite{grace-ad}  and BGRL~\cite{bgrl}, and highlight the merits of CCA-SSG. A quick overview is presented in Table~\ref{tbl-comparison}.

\textbf{No reliance on negative samples}. Most of previous works highly rely on negative pairs to avoid collapse or interchangeable, trivial/degenerated solutions~\cite{dgi, mvgrl, grace, grace-ad}. E.g., DGI and MVGRL generate negative examples by corrupting the graph structure severely, and GRACE/GCA treats all the other nodes within a graph as negative examples. However, for self-supervised learning on graphs, it is non-trivial to construct informative negative examples since nodes are structurally connected, and selecting negative examples in an arbitrary manner may lead to large variance for stochastic gradients and slow training convergence \cite{negative-flaw1}. The recently proposed BGRL model adopts asymmetric encoder architectures for SSL on graphs without the use of negative samples. However, though BGRL could avoid collapse empirically, it still remains as an open problem concerning its theoretical guarantee for preventing trivial solutions~\cite{tian-byol}. Compared with these methods, our model does not rely on negative pairs and asymmetric encoders. The feature decorrelation term can naturally prevent trivial solutions caused by the invariance term. We discuss the collapse issue detailedly in Appendix~\ref{append:collapse}. 

\textbf{No MI estimator, projector network nor asymmetric architectures}. Most previous works rely on additional components besides the GNN encoder to estimate some score functions in final objectives. DGI and MVGRL require a parameterized estimator to approximate mutual information between two views, and GRACE leverages a MLP projector followed by an InfoNCE estimator. BGRL harnesses asymmetric encoder architecture which consists of EMA (Exponential Moving Average), Stop-Gradient and an additional projector. MVGRL also induces asymmetric architectures as it adopts two different GNNs for the input graph and the diffusion graph respectively. In contrast, our approach requires no additional components except a single GNN encoder.

\textbf{Better efficiency and scalability to large graphs}. Consider a graph with $N$ nodes. DGI and MVGRL contrast node embeddings with graph embedding, which would require $O(N)$ space cost. GRACE treats two views of the same node as positive pairs and treat views of different nodes as negative pairs, which would take $O(N^2)$ space. BGRL focuses only on positive pairs, which will also take $O(N)$ space. By contrast, our method works on feature dimension. If we embed each node into a $D$-dimensional vector, the computation of the loss function would require $O(D^2)$ space. This indicates that the memory cost does not grow consistently as the size of graph increases. As a result, our method is promising for handling large-scale graphs without prohibitively large space costs.

\section{Theoretical Insights with Connection to  Information Theory}\label{sec:insights}

In this section we provide some analysis of the proposed objective function: 1) Interpretation of the loss function with entropy and mutual information. 2) The connection between the proposed objective and the Information Bottleneck principle. 3) Why the learned representations would be informative to downstream tasks. The proofs of propositions, theorems and corollaries are in Appendix~\ref{append:proof}.

\textbf{Notations.} Denote the random variable of input data as $X$ and the downstream task as $T$ (it could be the label $Y$ if the downstream task is classification). Note that in SSL, we have no access to $T$ in training and here we introduce the notation for our analysis. Define $S$ as the self-supervised signal (i.e., an augmented view of $X$), and $S$ shares the same space as $X$. Our model learns a representation for the input, denoted by $Z_X$ and its views, denoted by $Z_S$. $Z_X = f_{\theta}(X), Z_S = f_{\theta}(S)$, $f_{\theta}(\cdot)$ is a encoder shared by the original data and its views, which is parameterized by $\theta$. The target of representation learning is to learn a optimal encoder parameter $\theta$. Furthermore, for random variable $A, B, C$, we use $I(A,B)$ to denote the mutual information between $A$ and $B$, $I(A,B|C)$ to denote conditional mutual information of $A$ and $B$ on a given $C$, $H(A)$ for the entropy, and $H(A|B)$ for conditional entropy. The proofs of propositions, theorems and corollaries are in Appendix~\ref{append:proof}.

\subsection{An Entropy and Mutual Information Interpretation of the Objective}\label{sec-thm-1}
We first introduce an assumption about the distributions of $P(Z_S)$ and $P(Z_S|X)$. 
\begin{assumption}\label{assumption:gaussian}
    (Gaussian assumption of $P(Z_S|X)$ and $P(Z_S)$):
\begin{equation}
     P(Z_S|X) = \mathcal{N} (\mu_X, \Sigma_X), P(Z_S) = \mathcal{N}(\mu, \Sigma).
\end{equation}
\end{assumption}
With Assumption~\ref{assumption:gaussian}, we can arrive at the following propositions:
\begin{proposition}\label{prop:conditional-entropy}
   In expectation, minimizing Eq.~\eqref{eqn:loss-inv} is equivalent to minimizing the entropy of $Z_{S}$ conditioned on input $X$, i.e.,
    \begin{equation}
        \min\limits_{\theta} \mathcal{L}_{inv} \cong \min\limits_{\theta} H(Z_{S} | X).
    \end{equation}
\end{proposition}
\begin{proposition}\label{prop:entropy}
    Minimizing Eq.~\eqref{eqn:loss-decor} is equivalent to maximizing the entropy of $Z_S$, i.e.,
    \begin{equation}
        \min\limits_{\theta} \mathcal{L}_{dec} \cong \max\limits_{\theta} H(Z_{S}).
    \end{equation}
\end{proposition}
The two propositions unveil the effects of two terms in our objective. Combining two propositions, we can further interpret Eq.~\eqref{eqn:loss} from an information-theoretic perspective.
\begin{theorem}\label{the:mi}
    By optimizing Eq~\eqref{eqn:loss}, we maximize the mutual information between the augmented view's embedding $Z_S$ and the input data $X$, and minimize the mutual information between $Z_S$ and the view itself $S$, conditioned on the input data $X$. Formally we have
   \begin{equation}
      \min\limits_{\theta} \mathcal{L} \Rightarrow \max\limits_{\theta} I(Z_S, X) \;\mbox{and}\; \min\limits_{\theta} I(Z_S, S|X).
   \end{equation}
\end{theorem}
The proof is based on the facts $I(Z_S, X) = H(Z_S) - H(Z_S|X)$ and $I(Z_S, S|X) = H(Z_S|X) + H(Z_S|S) = H(Z_S|X)$. Theorem~\ref{the:mi} indicates that our objective~Eq.~\eqref{eqn:loss} learns representations that maximize the information of the input data, i.e., $I(Z_S, X)$, and meanwhile minimize the lost information during augmentation, i.e., $I(Z_S, S|X)$.

\subsection{Connection with the Information Bottleneck Principle}
The analysis in Section~\ref{sec-thm-1} enables us to further build a connection between our objective Eq.~\eqref{eqn:loss} and the well-studied Information Bottleneck Principle~\cite{IB-1,IB-2,IB-3,IB-4} under SSL settings. Recall that the supervised Information Bottleneck (IB) is defined as follows:
\begin{definition}
The supervised IB aims at maximizing an Information Bottleneck Lagrangian:
\begin{equation}\label{eqn:ib}
    \mathcal{IB}_{sup} = I(Y, Z_X) - \beta I(X, Z_X), \;\mbox{where}\; \beta > 0.
\end{equation}
\end{definition}
As we can see, $\mathcal{IB}_{sup}$ attempts to maximize the information between the data representation $Z_X$ and its corresponding label $Y$, and concurrently minimize the information between $Z_X$ and the input data $X$ (i.e., exploiting compression of $Z_X$ from $X$). The intuition of IB principle is that $Z_X$ is expected to contain only the information that is useful for predicting $Y$.

Several recent works~\cite{Multi-IB, mini-1, barlow} propose various forms of IB under self-supervised settings. The most relevant one names Self-supervised Information Bottleneck:
\begin{definition}\label{def:ibssl}
   (Self-supervised Information Bottleneck~\cite{barlow}). The Self-supervised IB aims at maximizing the following Lagrangian: \begin{equation}\label{eqn:ib-ssl}
       \mathcal{IB}_{ssl} = I(X, Z_{S}) - \beta I(S, Z_{S}), \;\mbox{where}\; \beta > 0.
   \end{equation}
\end{definition}
Intuitively, $\mathcal{IB}_{ssl}$ posits that a desirable representation is expected to be informative to augmentation invariant features, and to be a maximally compressed representation of the input.

Our objective Eq.~\eqref{eqn:loss} is essentially an embodiment of $\mathcal{IB}_{ssl}$: 
\begin{theorem}\label{the:ibssl}
Assume $0 < \beta \le 1$, then by minimizing Eq.~\eqref{eqn:loss}, the self-supervised Information Bottleneck objective is maximized, formally:
   \begin{equation}
       \min\limits_{\theta} \mathcal{L} \Rightarrow \max\limits_{\theta} \mathcal{IB}_{ssl}
   \end{equation}
   
\end{theorem}
 Theorem~\ref{the:ibssl} also shows that Eq.~\eqref{eqn:loss} implicitly follows the same spirit of IB principle under self-supervised settings. As further enlightenment, we can relate Eq.~\eqref{eqn:loss} with the \emph{multi-view Information Bottleneck}~\cite{Multi-IB} and the \emph{minimal and sufficient representations for self-supervision}~\cite{mini-1}:
\begin{corollary}\label{corollary:1}
    Let $X_1 = S$, $X_2 = X$ and assume $0< \beta \le 1$, then minimizing Eq.~\eqref{eqn:loss} is equivalent to minimizing the Multi-view Information Bottleneck Loss in~\cite{Multi-IB}:
    \begin{equation}\label{eqn:mib}
       \mathcal{L}_{MIB} = I(Z_1, X_1 | X_2) - \beta I(X_2, Z_1), \;\mbox{where}\; 0 <\beta \le 1.
    \end{equation}
\end{corollary}
\begin{corollary}\label{corollary:2}
     When the data augmentation process is reversible, minimizing Eq.~\eqref{eqn:loss} is equivalent to learning the Minimal and Sufficient Representations for Self-supervision in~\cite{mini-1}:
     \begin{equation}\label{eqn:mini}
        Z^{\text{ssl}}_{X} = \mathop{\arg \max}\limits_{Z_X} I(Z_X, S), Z^{\text{ssl}_{\text{min}}}_X = \mathop{\arg \min}\limits_{Z_X} H(Z_X | S) \;\; \mbox{s.t.} \;\; I(Z_X, S) \text{ is maximized}.
    \end{equation}
\end{corollary}

\subsection{Influence on Downstream Tasks}\label{sec:analysis-downstream}
We have provided a principled understanding for our new objective. Next, we discuss its effect on downstream tasks $T$. The rationality of data augmentations in SSL is rooted in a conjecture that an ideal data augmentation approach would not change the information related to its label. We formulate this hypothesis as a building block for analysis on downstream tasks~\cite{info-multi, Multi-IB}.
\begin{assumption}\label{assumption:redundancy}
    (Task-relevant information and data augmentation).
    All the task-relevant information is shared across the input data $X$ and its augmentations $S$, i.e., $I(X,T) = I(S,T) = I(X,S,T)$, or equivalently,
    $I(X,T|S) = I(S,T|X) = 0 $.
\end{assumption}

This indicates that all the task-relevant information is contained in augmentation invariant features. We proceed to derive the following theorem which reveals the efficacy of the learned representations by our objective with respect to downstream tasks.
\begin{theorem}\label{the:downstream}
   (Task-relevant/irrelevant information). By optimizing Eq.~\eqref{eqn:loss}, the task-relevant information $I(Z_S, T)$ is maximized, and the task-irrelevant information $H(Z_S|T)$ is minimized. Formally,
  \begin{equation}
      \min\limits_{\theta}\mathcal{L} \Rightarrow \max\limits_{\theta} I(Z_S,T) \;\mbox{and}\;\min\limits_{\theta} H(Z_S | T).
  \end{equation}
\end{theorem}
Therefore, the learned representation $Z_S$ is expected to contain minimal and sufficient information about downstream tasks~\cite{mini-1, Multi-IB}, which further illuminates the reason why the embeddings given by SSL approaches have superior performance on various downstream tasks.

\section{Experiments} \label{sec:experiments}
We assess the quality of representations after self-supervised pretraining on seven node classification benchmarks: \textit{Cora, Citeseer, Pubmed, Coauthor CS, Coauthor Physics} and \textit{Amazon Computer, Amazon-Photo}. We adopt the public splits for \textit{Cora, Citeseer, Pubmed}, and a 1:1:9 training/validation/testing splits for the other 4 datasets. Details of the datasets are in Appendix~\ref{append:exp}.

\textbf{Evaluation protocol}. We follow the linear evaluation scheme as introduced in~\cite{dgi}: \textbf{i)} We first train the model on all the nodes in a graph without supervision, by optimizing the objective in Eq.~\eqref{eqn:loss}. \textbf{ii)} After that, we freeze the parameters of the encoder and obtain all the nodes' embeddings, which are subsequently fed into a linear classifier (i.e., a logistic regression model) to generate a predicted label for each node. In the second stage, only nodes in training set are used for training the classifier, and we report the classification accuracy on testing nodes. 

We implement the model with PyTorch. All experiments are conducted on a NVIDIA V100 GPU with 16 GB memory. We use the Adam optimizer~\cite{adam} for both stages. The graph encoder $f_{\theta}$ is specified as a standard two-layer GCN model \cite{gcn} for all the datasets except \textit{citeseer} (where we empirically find that a one-layer GCN is better). We report the mean accuracy with a standard deviation through $20$ random initialization (on \textit{Coauthor CS, Coauthor Physics} and \textit{Amazon Computer, Amazon-Photo}, the split is also randomly generated). Detailed hyperparameter settings are in Appendix~\ref{append:exp}.

\subsection{Comparison with Peer Methods}
We compare CCA-SSG with classical unsupervised models, Deepwalk~\cite{deepwalk} and GAE~\cite{gae}, and self-supervised models, DGI~\cite{dgi}, MVGRL~\cite{mvgrl}, GRACE~\cite{grace} and GCA~\cite{grace-ad}. We also compare with supervised learning models, including MLP, Label Propagation (LP)~\cite{lp}, and supervised baselines GCN~\cite{gcn} and GAT~\cite{gat}\footnote{The BGRL~\cite{bgrl} is not compared as its source code has not been released.}. The results of baselines are quoted from ~\cite{mvgrl,grace, grace-ad} if not specified.
\begin{table}[tb!]
	\centering
	\caption{Test accuracy on citation networks. The \emph{input} column highlights the data used for training. ($\mathbf{X}$ for node features, $\mathbf{A}$ for adjacency matrix, $\mathbf{S}$ for diffusion matrix, and $\mathbf{Y}$ for node labels). }
	\label{tbl-exp-citation}
	\small
	\begin{threeparttable}
    {\setlength{\tabcolsep}{4mm}{
		\begin{tabular}{c|lcccc}
			\toprule[0.8pt]
		        &	Methods & Input & Cora & Citeseer & Pubmed \\
			\midrule
		        \multirow{4}{*}{\rotatebox{0}{Supervised}} 
		            & MLP \cite{gat}  & $\mathbf{X, Y}$  & 55.1 & 46.5 & 71.4 \\
                    & LP \cite{lp} 	& $\mathbf{A, Y}$ & 68.0 & 45.3 & 63.0 \\
                    & GCN \cite{gcn}  & $\mathbf{X, A, Y}$ & 81.5 & 70.3 & 79.0 \\
                    & GAT \cite{gat}   & $\mathbf{X, A, Y}$ & 83.0 $\pm$ 0.7  & 72.5 $\pm$ 0.7 & 79.0 $\pm$ 0.3			                                  \\
            \midrule[0.5pt]
            	\multirow{8}{*}{\rotatebox{0}{Unsupervised}} 
            	    & Raw Features~\cite{dgi}  &  $\mathbf{X}$ & 47.9 $\pm$ 0.4  & 49.3 $\pm$ 0.2 & 69.1 $\pm$ 0.3	            \\
            	    & Linear CCA~\cite{cca}   &  $\mathbf{X}$ & 58.9 $\pm$ 1.5   & 27.5 $\pm$ 1.3 &75.8 $\pm$ 0.4  \\ 
                    & DeepWalk~\cite{deepwalk} &	 $\mathbf{A}$ &  70.7 $\pm$ 0.6  & 51.4 $\pm$ 0.5 & 74.3 $\pm$ 0.9 \\
                    & GAE~\cite{gae}      &  $\mathbf{X, A}$ & 71.5 $\pm$ 0.4  & 65.8 $\pm$ 0.4 & 72.1 $\pm$ 0.5	\\
                    & DGI~\cite{dgi}     & $\mathbf{X, A}$ & 82.3 $\pm$ 0.6  & 71.8 $\pm$ 0.7 & 76.8 $\pm$ 0.6 \\
                    & MVGRL$^{1}$~\cite{mvgrl}    & $\mathbf{X, S, A}$ & 83.5 $\pm$ 0.4  & \textbf{73.3} $\pm$ \textbf{0.5} & 80.1 $\pm$ 0.7\\
                    & GRACE$^2$~\cite{grace}  & $\mathbf{X, A}$ & 81.9 $\pm$ 0.4  & 71.2 $\pm$ 0.5 & 80.6 $\pm$ 0.4	     		                                  \\
                    & CCA-SSG (Ours)    & $\mathbf{X, A}$ &  \textbf{84.2} $\pm$ \textbf{0.4}   & 73.1 $\pm$ 0.3 &  \textbf{81.6} $\pm$ \textbf{0.4}\\
			\bottomrule[0.8pt]
		\end{tabular}}
		}
		\begin{tablenotes}
		\item[1] Results on Cora with authors' code is inconsistent with~\cite{mvgrl}. We adopt the results with authors' code. 
        \item[2] Results are from our reproducing with authors' code, as~\cite{grace} did not use the public splits.
        \end{tablenotes}
	\end{threeparttable}
\end{table}
\begin{table}[tb!]
	\centering
	\caption{Test accuracy on co-author and co-purchase networks. We report both mean accuracy and standard deviation. Results of baseline models are from ~\cite{grace-ad}.}
	\label{tbl-exp-co}
	\small
	\begin{threeparttable}
        {
		\begin{tabular}{c|lccccc}
			\toprule[0.8pt]
		        &	Methods & Input & Computer & Photo & CS & Physics \\
			\midrule
		          &Supervised GCN \cite{gcn}   & $\mathbf{X, A, Y}$ & 86.51 $\pm$ 0.54 & 92.42 $\pm$ 0.22 & 93.03 $\pm$ 0.31 & 95.65 $\pm$ 0.16 \\
                   &Supervised GAT \cite{gat}  & $\mathbf{X, A, Y}$ & 86.93 $\pm$ 0.29 & 92.56 $\pm$ 0.35 & 92.31 $\pm$ 0.24 & 95.47 $\pm$ 0.15 \\
                  \midrule[0.5pt]
            \multirow{8}{*}{\rotatebox{90}{Unsupervised}} 
                & Raw Features~\cite{dgi}  &  $\mathbf{X}$ & 73.81 $\pm$ 0.00  & 78.53 $\pm$ 0.00 & 90.37 $\pm$ 0.00 & 93.58 $\pm$ 0.00	            \\
                           	    & Linear CCA~\cite{cca} &  $\mathbf{X}$ & 79.84 $\pm$ 0.53   & 86.92 $\pm$ 0.72 &93.13 $\pm$ 0.18  & 95.04 $\pm$ 0.17    \\ 
                    & DeepWalk \cite{deepwalk} &	 $\mathbf{A}$ &  85.68 $\pm$ 0.06  & 89.44 $\pm$ 0.11 & 84.61 $\pm$ 0.22 & 91.77 $\pm$ 0.15 \\
                    & DeepWalk + features   &	 $\mathbf{X, A}$ &  86.28 $\pm$ 0.07  & 90.05 $\pm$ 0.08 & 87.70 $\pm$ 0.04  & 94.90 $\pm$ 0.09\\
                    & GAE \cite{gae}   &  $\mathbf{X, A}$ & 85.27 $\pm$ 0.19  & 91.62 $\pm$ 0.13 & 90.01 $\pm$ 0.71 & 94.92 $\pm$ 0.07		\\
                    & DGI  \cite{dgi}    & $\mathbf{X, A}$ & 83.95 $\pm$ 0.47  & 91.61 $\pm$ 0.22 & 92.15 $\pm$ 0.63 & 94.51 $\pm$ 0.52\\
                    & MVGRL \cite{mvgrl}   & $\mathbf{X, S, A}$ & 87.52 $\pm$ 0.11  & 91.74 $\pm$ 0.07 & 92.11 $\pm$ 0.12 & 95.33 $\pm$ 0.03  \\
                    & GRACE$^{1}$ \cite{grace}  & $\mathbf{X, A}$ & 86.25 $\mathbf{\pm}$ 0.25  & 92.15 $\pm$ 0.24 & 92.93 $\pm$ 0.01 & 95.26 $\pm$ 0.02	  \\
                    
                    & GCA$^{1}$ \cite{grace-ad} & $\mathbf{X, A}$ & 87.85 $\pm$ 0.31  & 92.49 $\pm$ 0.09 & 93.10 $\pm$ 0.01 & \textbf{95.68} $\pm$ \textbf{0.05}	  \\
                      
                    & CCA-SSG (Ours)    & $\mathbf{X, A}$ & \textbf{88.74} $\pm$ \textbf{0.28} & \textbf{93.14} $\pm$ \textbf{0.14} & \textbf{93.31} $\pm$ \textbf{0.22} & 95.38 $\pm$ 0.06\\
			\bottomrule[0.8pt]
		\end{tabular}}
				\begin{tablenotes}
        \item[1] GCA is essentially an enhanced version of GRACE by adopting adaptive augmentations. Both GRACE and GCA would suffer from \emph{out of memory} on \textit{Coauthor-Physics} using a GPU wth 16GB memory. The reported results are from authors' papers using a 32GB GPU.
        \end{tablenotes}
	\end{threeparttable}
\end{table}

We report the node classification results of citation networks and other datasets in Table~\ref{tbl-exp-citation} and Table~\ref{tbl-exp-co} respectively. As we can see, CCA-SSG outperforms both the unsupervised competitors and the fully supervised baselines on \textit{Cora} and \textit{Pubmed}, despite its simple architecture. On \textit{Citeseer}, CCA-SSG achieves competitive results as of the most powerful baseline MVGRL. On four larger benchmarks, CCA-SSG also achieves the best performance in four datasets except \textit{Coauther-Physics}. It is worth mentioning that we empirically find that on \textit{Coauthor-CS} a pure 2-layer-MLP encoder is better than GNN models. This might because the graph-structured information is much less informative than the node features, presumably providing harmful signals for classification (in fact, on \textit{Coauthor-CS}, linear models using merely node features can greatly outperform DeepWalk/DeepWalk+features).

\subsection{Ablation Study and Scalability Comparison}
\begin{figure}[t!]
    \begin{minipage}{0.35\linewidth}
    \centering
	\captionof{table}{Ablation study of node classification accuracy (\%) on the key components of CCA-SSG.}
	\label{tbl-exp-abl}
	\small
    \setlength{\tabcolsep}{0.8mm}
    { 
	\begin{threeparttable}
    {
		\begin{tabular}{cccc}
			\toprule[0.8pt]
		    Variants  & {Cora} & {Citeseer} & {Pubmed} \\
		    \midrule[0.5pt]
		    Baseline  & 84.2  & 73.1 & 81.6  \\
		    \midrule[0.5pt]
			w/o $\mathcal{L}_{dec}$ & 79.1 & 72.2 & 75.3  \\
			w/o $\mathcal{L}_{inv}$ & 40.1 & 28.9 & 46.5 \\
			\bottomrule[0.8pt]
		\end{tabular}}
	\end{threeparttable}
	}
    \end{minipage}
    \begin{minipage}{0.32\linewidth}
    \centering
    \includegraphics[width=1\textwidth]{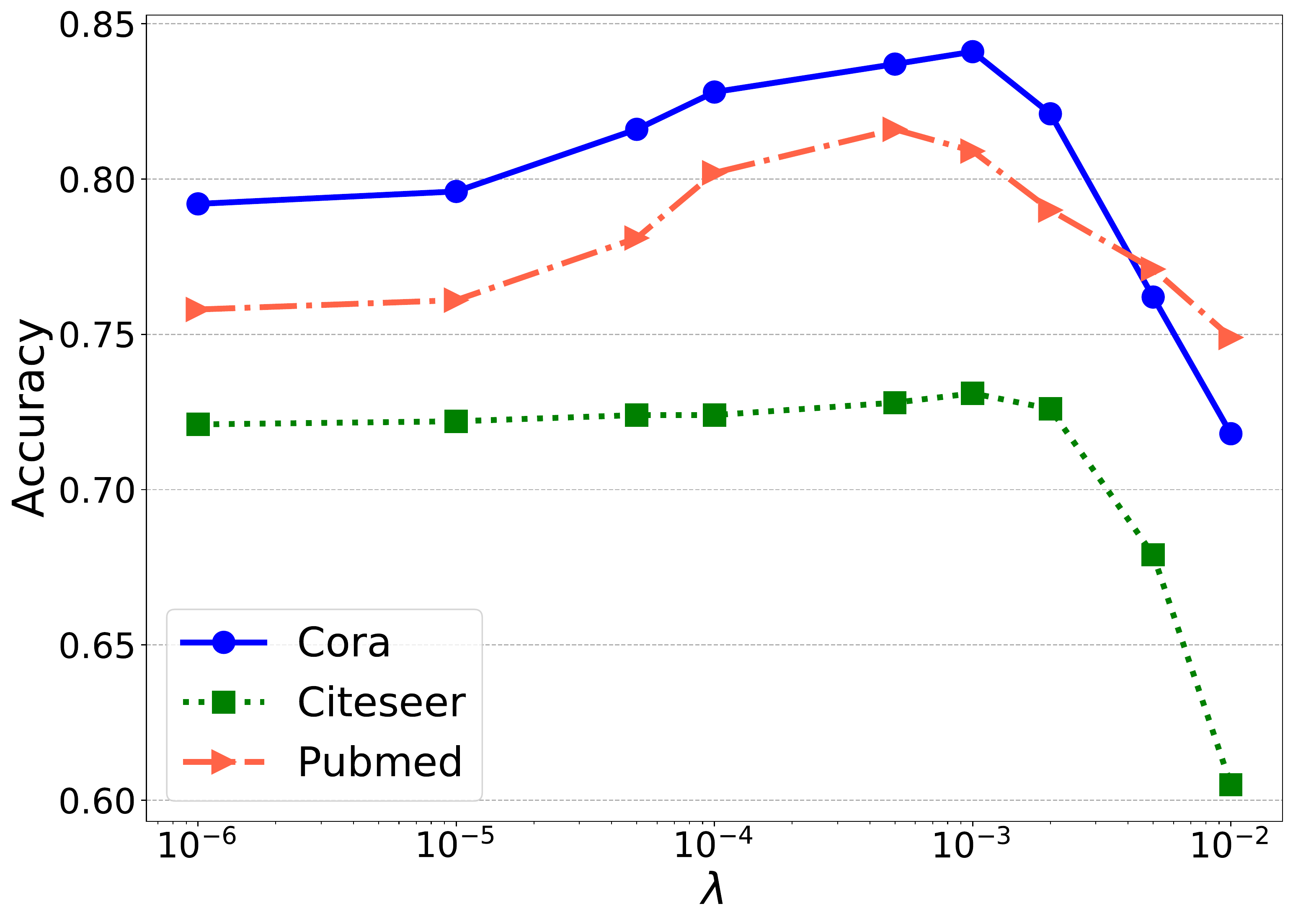}
    \caption{Effect of $\lambda$.}
    \label{fig:abl-lam}
    \end{minipage}
    \begin{minipage}{0.32\linewidth}
    \centering
    \includegraphics[width=1\textwidth]{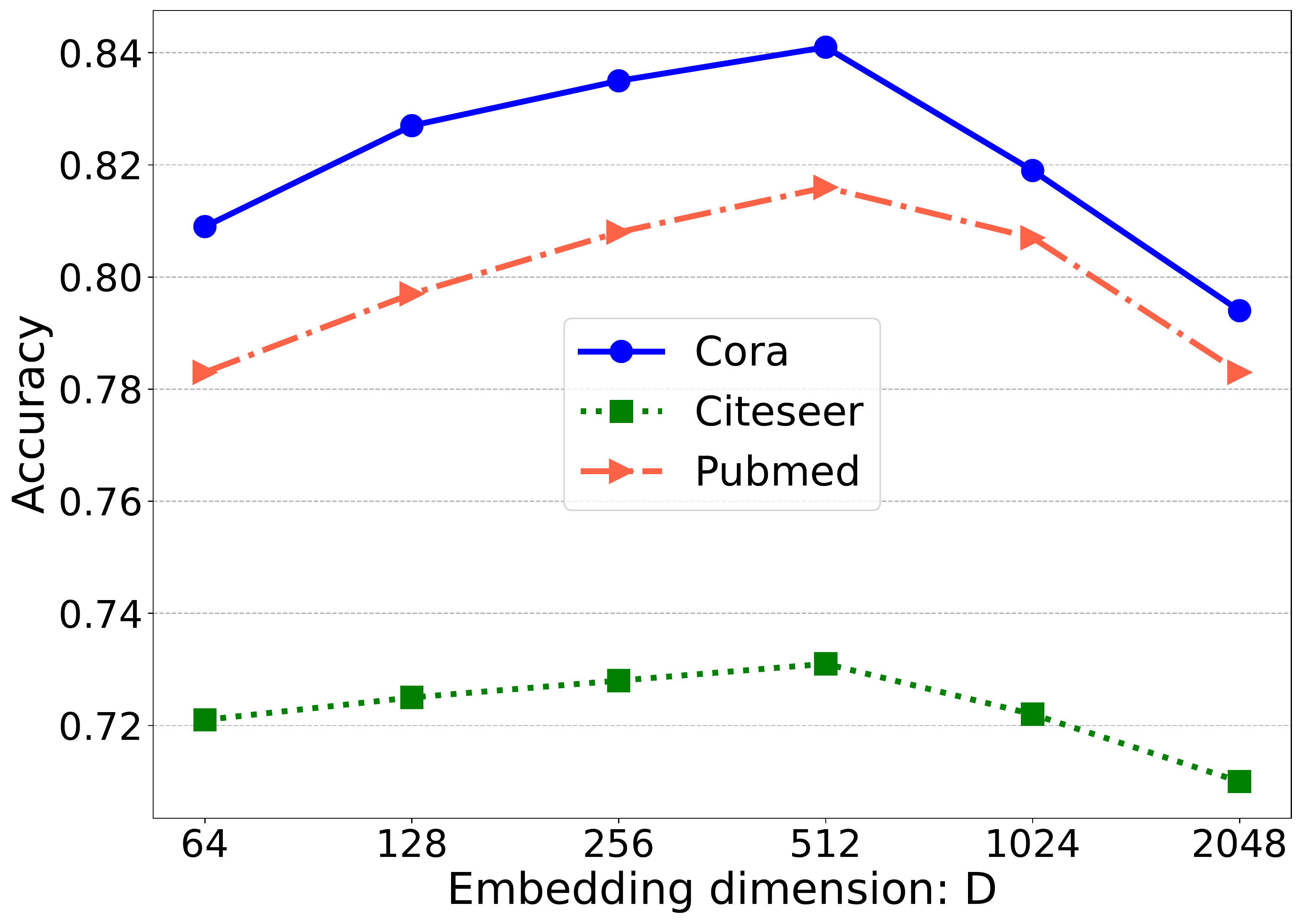}
    \caption{Effect of $D$.}
    \label{fig:abl-D}
    \end{minipage}
 
\end{figure}

\textbf{Effectiveness of invariance/decorrelation terms}. We alter our loss by removing the invariance/decorrelation term respectively to study the effects of each component, with results reported in Table~\ref{tbl-exp-abl}. We find that only using the invariance term will lead to merely performance drop instead of completely collapsed solutions. This is because node embeddings are normalized along the instance dimension to have a zero-mean and fixed-standard deviation, and the worst solution is no worse than dimensional collapse (i.e., all the embeddings lie in an line, and our decorrelation term can help to prevent it) instead of complete collapse (i.e., all the embeddings degenerate into a single point). As expected, only optimizing the decorrelation term will lead to poor result, as the model learns nothing meaningful but disentangled representation. In Appendix~\ref{append:collapse} we discuss the relationship between complete/dimensional collapse, when the two cases happen and how to avoid them.

\textbf{Effect of decorrelation intensity}. We study how the intensity of feature decorrelation improves/degrades the performance by increasing the trade-off hyper-parameter $\lambda$. Fig.~\ref{fig:abl-lam} shows test accuracy w.r.t. different $\lambda$'s on \textit{Cora, Citeseer} and \textit{Pubmed}. The performance benefits from a proper selection of $\lambda$ (from $0.0005$ to $0.001$ in our experiments). When $\lambda$ is too small, the decorrelation term does not work; if it is too large, the invariance term would be neglected, leading to serious performance degrade. An interesting finding is that even when $\lambda$ is very small or even equals to $0$ (w/o $\mathcal{L}_{dec}$ in Table~\ref{tbl-exp-abl}), the test accuracy on \textit{Citeseer} does not degrade as much as that on \textit{Cora} and \textit{Citeseer}. The reason is that node embeddings of \textit{Citeseer} is already highly uncorrelated even without the decorrelation term. Appendix~\ref{append:add} visualizes the correlation matrices without/with decorrelations. 

\textbf{Effect of embedding dimension}. Fig.~\ref{fig:abl-D} shows the effect of the embedding dimension. Similar to contrastive methods~\cite{dgi, mvgrl, grace, grace-ad}, CCA-SSG benefits from a large embedding dimension (compared with supervised learning), while the optimal embedding dimension of CCA-SSG ($512$ on most benchmarks) is a bit larger than other methods (usually $128$ or $256$). Yet, we notice a performance drop as the embedding dimension  increases. We conjecture that the CCA is essentially a dimension-reduction method, the ideal embedding dimension ought to be smaller than the dimension of input. Hence we do not apply it on well-compressed datasets (e.g. ogbn-arXiv and ogbn-product).

\textbf{Scalability Comparison.}
Table~\ref{exp-tbl-scale} compares model size, training time (till the epoch that gives the highest evaluation accuracy) and memory cost of CCA-SSG with other methods, on \textit{Cora, Pubmed} and \textit{Amazon-Computers}. Overall, our method has fewer parameters, shorter training time, and fewer memory cost than MVGRL, GRACE and GCA in most cases. DGI is another simple and efficient model, but it yields much poorer performance. The results show that despite its simplicity and efficiency, our method achieves even better (or competitive) performance.

\begin{table}[tb!]
	\centering
	\caption{Comparison of the number of parameters, training time for achieving the best performance, and the memory cost of different methods on \textit{Cora, Pubmed} and \textit{Amazon-Computer}. MVGRL on Pubmed and Computer requires subgraph sampling with graph size $4000$. Others are full-graph.}
	\small
	\label{exp-tbl-scale}
	\begin{threeparttable}
        {
		\begin{tabular}{lccccccccc}
			\toprule[0.8pt]
					       	\multirow{2}{*}{{Methods}}   & \multicolumn{3}{c}{Cora ($N$: 2,708)} & \multicolumn{3}{c}{Pubmed ($N$: 19,717)} & \multicolumn{3}{c}{Computer ($N$: 13,752)} \\
		        \cline{2-4} \cline{5-7} \cline{8-10}
		             & \#Paras & Time & Mem  & \#Paras & Time & Mem  & \#Paras & Time & Mem\\
			\midrule
                     DGI & 1260K & 6.4s  & 1.4G & 782K  & 5.9s & 1.9G & 919K  &  14.1s & 1.9G  \\
                     MVGRL  & 1731K & 26.9s & 4.6G & 775K & 29s &  5.4G & 1049K & 31.5s & 5.5G \\
                     GRACE/GCA  & 997K & 8.3s & 1.7G & 520K & 756s & 12.6G  & 273K & 314s & 7.6G \\
                    CCA-SSG(Ours) & 997K & {3.8s} &  {1.6G} & 519K & {9.6s} & {2.7G} & 656K & {14.8s} & {2.5G} \\
			\bottomrule[0.8pt]
		\end{tabular}}
	\end{threeparttable}
\end{table}
\section{Conclusion and Discussions}\label{sec:discussion}
In this paper, we have introduced CCA-SSG, a conceptually simple,  efficient yet effective method for self-supervised representation learning on graphs, based on the idea of Canonical Correlation Analysis. Compared with contrastive methods, our model does not require  additional components except random augmentations and a GNN encoder, whose effectiveness is justified in experiments.
\paragraph{Limitations of the work.}\label{sec:discussion-limitation}
Despite the theoretical grounds and the promising experimental justifications, our method would suffer from several limitations. 1) The objective Eq.~\eqref{eqn:loss} is essentially performing dimension reduction, while SSL approach usually requires a large embedding dimension. As a result, our method might not work well on datasets where input data does not have a large feature dimension. 2) Like other augmentation based methods, CCA-SSG highly relies on a high-quality, informative and especially, label-invariant augmentations. However, the augmentations used in our model might not perfectly meet these requirements, and it remains an open problem how to generate informative graph augmentations that have non-negative impacts on the downstream tasks. 

\paragraph{Potential negative societal impacts.} \label{sec:discussion-negative}
This work explores a simple pipeline for representation learning without large amount of labeled data. However, in industry there are many career workers whose responsibility is to label or annotate data. The proposed method might reduce the need for labeling data manually, and thus makes a few individuals unemployed (especially for developing countries and remote areas). Furthermore, our model might be biased, as it tends to pay more attention to the majority and dominant features (shared information across most of the data). The minority group whose features are scare are likely to be downplayed by the algorithm.

\begin{ack}
This work was supported in part by NSF under grants III-1763325, III-1909323,  III-2106758, and SaTC-1930941. Qitian Wu and Junchi Yan were partly supported by Shanghai Municipal Science and Technology Major Project (2021SHZDZX0102). We thank Amazon Web Services for sponsoring computation resources for this work.
\end{ack}
{
\bibliographystyle{plain}
\bibliography{ref}
}
\newpage
\appendix

\section{Algorithm}\label{append:algorithm}
We provide the pseudo code for our method in Algorithm~\ref{alg:cca-ssg}, the detailed description of which is in Section~\ref{sec:framework}.
\begin{algorithm}[htbp]
 \caption{Algorithm for CCA-SSG}
 \label{alg:cca-ssg}
 \KwIn{A graph $\mathbf{G} = (\mathbf{X}, \mathbf{A})$  with $N$ nodes, where $\mathbf{X}$ is node feature matrix, and $\mathbf{A}$ is the adjacency matrix. Augmentations $\mathcal{T}$, encoder $f_{\theta}$ by random initialization, trade-off hyper-parameter $\lambda$, maximum training steps ${T}$.}
 \KwOut{Learned encoder $f_{\theta}$.}
 \BlankLine
  \While{not reaching $T$}{
    Sample two augmentation functions $t_A, t_B \sim \mathcal{T}$\;
    Generate transformed graphs: $\tilde{\mathbf{G}}_A = (\tilde{\mathbf{X}}_A, \tilde{\mathbf{A}}_A)$, $\tilde{\mathbf{G}}_B = (\tilde{\mathbf{X}}_B, \tilde{\mathbf{A}}_B)$ \;
    Get node embeddings through the graph neural network as encoder: $\bm{Z}_A = f_{\theta}(\tilde{\mathbf{X}}_A, \tilde{\mathbf{A}}_A), \bm{Z}_B = f_{\theta}(\tilde{\mathbf{X}}_B, \tilde{\mathbf{A}}_B) $ \;
    Normalize embeddings along instance dimension:
    $\tilde{\mathbf{Z}}_A = \frac{\mathbf{Z}_A - \mu(\mathbf{Z}_A)}{\sigma(\mathbf{Z}_A)*\sqrt{N}}, \tilde{\mathbf{Z}}_B = \frac{\mathbf{Z}_B - \mu(\mathbf{Z}_B)}{\sigma(\mathbf{Z}_B)*\sqrt{N}}$ \;
    Calculate the loss function $\mathcal{L}$ according to Eq.~\eqref{eqn:loss} \;
    Update $\theta$ by gradient descent\; 
 }
\textbf{Inference}: $\mathbf{Z} = f_{\theta}(\mathbf{X}, \mathbf{A})$, where $\theta$ is the frozen parameters of the encoder.
\end{algorithm}
\section{Discussions on Degenerated Solutions in SSL}\label{append:collapse}
In this section we provide an illustration and some discussions for degenerated (collapsed) solutions, or namely trivial solutions, in self-supervised representation learning. The discussion is inspired by the separation of complete collapse and dimensional collapse proposed in~\cite{decorrelation}. We show that our method naturally avoids complete collapse through feature-wise normalization, and could prevent/alleviate dimensional collapse through the decorrelation term Eq.~\eqref{eqn:loss-decor}.

In most contrastive learning methods especially the augmentation-based ones~\cite{cpc, moco, simclr, cmc}, both positive pairs and negative pairs are required for learning a model. For instance, the widely adopted InfoNCE~\cite{cpc} loss has the following formulation:
\begin{equation}\label{eqn:infonce}
    \mathcal{L}_{\text{InfoNCE}} = -\log \frac{\exp\left(z^A_i\cdot z^B_i / \tau\right)}{\sum_{j} \exp\left(z_i^A \cdot z_j^B / \tau\right)},
\end{equation}
where $z^A_i$ and $z^B_i$ are the (normalized) embeddings of two views of the same instance $i$, and $\tau$ is the temperature hyperparameter. The numerator enforces similarity between positive pairs (two views of the same instance), while the denominator promotes dis-similarity between negative pairs (two views of different instances). Therefore, minimizing Eq.~\eqref{eqn:infonce} is equivalent to maximizing the cosine similarity of positive pairs and meanwhile minimizing the cosine similarity of negative pairs. Note that the normalization is applied for each instance (projecting the embedding onto a hypersphere), so we are essentially minimizing the distances between positive pairs and maximizing the distance between negative pairs. The previous work~\cite{hypersphere} provides a thorough analysis on the behaviors of the objective by decomposing it into two terms: 1) alignment term (for positive pairs) and 2) uniformity term (for negative pairs). 

The alignment loss is defined as the expected distance between positive pairs:
\begin{equation}\label{eqn:align}
    \mathcal{L}_{\text{align}} \triangleq  \mathbb{E}_{(x,y)\sim p_{\text{pos}}} \left\| f(x) - f(y) \right\|_2^{\alpha}, \;\mbox{with}\; \alpha>0.
\end{equation}

The uniformity loss is the logarithm of the average pairwise Gaussian potential:
\begin{equation}\label{eqn:uniform}
    \mathcal{L}_{\text{uniformity}}  \triangleq  \log \mathbb{E}_{x,y\sim p_{\text{data}}} e^{-t\left\| f(x) - f(y)\right\|_2^2},  \;\mbox{with}\;  t>0
\end{equation}
Intuitively, the alignment term makes the positive pairs close to each other on the hypersphere, while the uniformity term makes different data points distribute on the hypersphere uniformly.

In particular, only considering the alignment term in Eq.~\eqref{eqn:align} will lead to trivial solutions: all the embeddings would degenerate to a fixed point on the hypersphere. This phenomenon is called \textbf{complete collapse} ~\cite{decorrelation}. Denote $\mathbf{Z}_A$ and $\mathbf{Z}_B$ as two embedding matrix of two views ($\mathbf{Z} \in \mathbb{R}^{N\times D}$ and is row normalized), then in this case $\mathbf{Z}_A\mathbf{Z}_B^\top \cong \bm{1}$ is an all-one matrix (so as $\mathbf{Z}_A\mathbf{Z}_A^\top$ and $\mathbf{Z}_B\mathbf{Z}_B^\top$).

\begin{figure}[tb!]
\centering
\subfigure[complete collapse]{
\begin{minipage}[t]{0.22\linewidth}
\centering
\includegraphics[width=1\textwidth,angle=0]{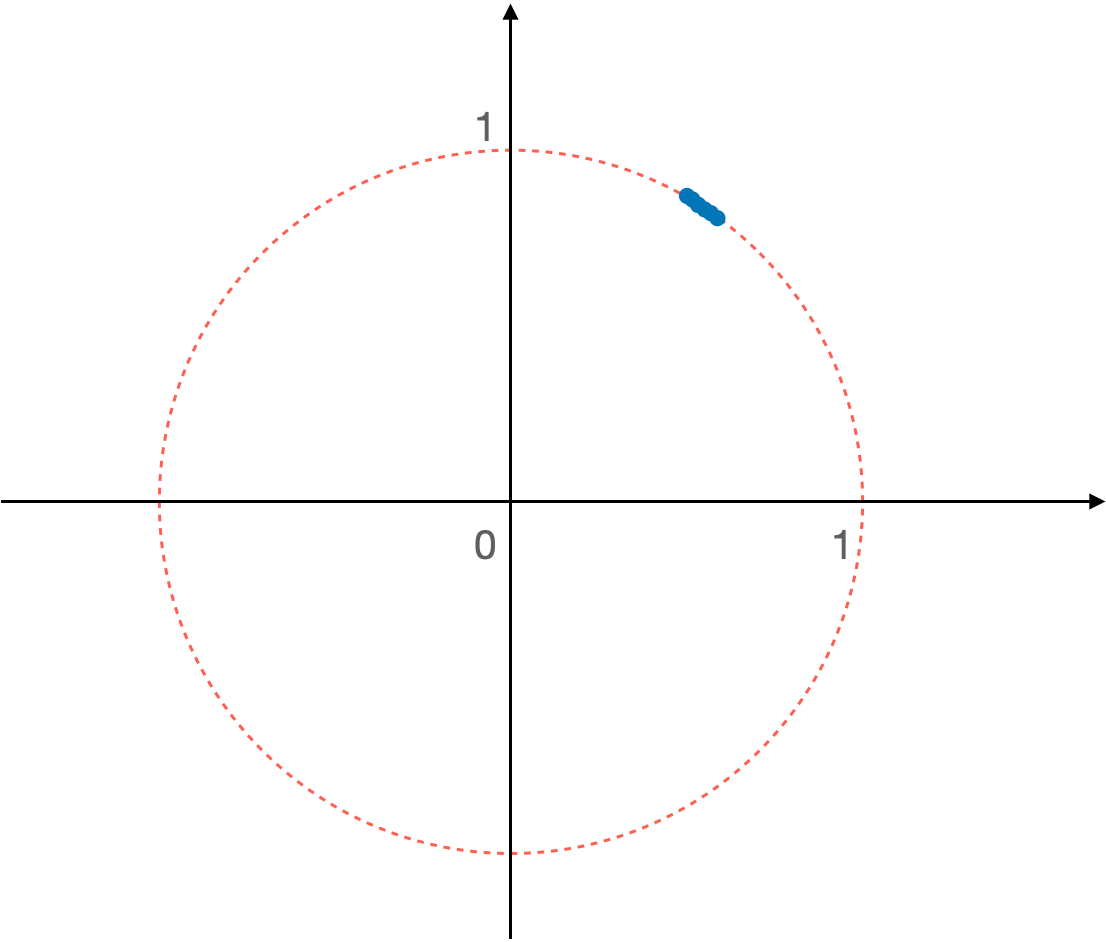}
\label{fig:complete}
\end{minipage}%
}%
\subfigure[uniformity]{
\begin{minipage}[t]{0.22\linewidth}
\centering
\includegraphics[width=1\textwidth,angle=0]{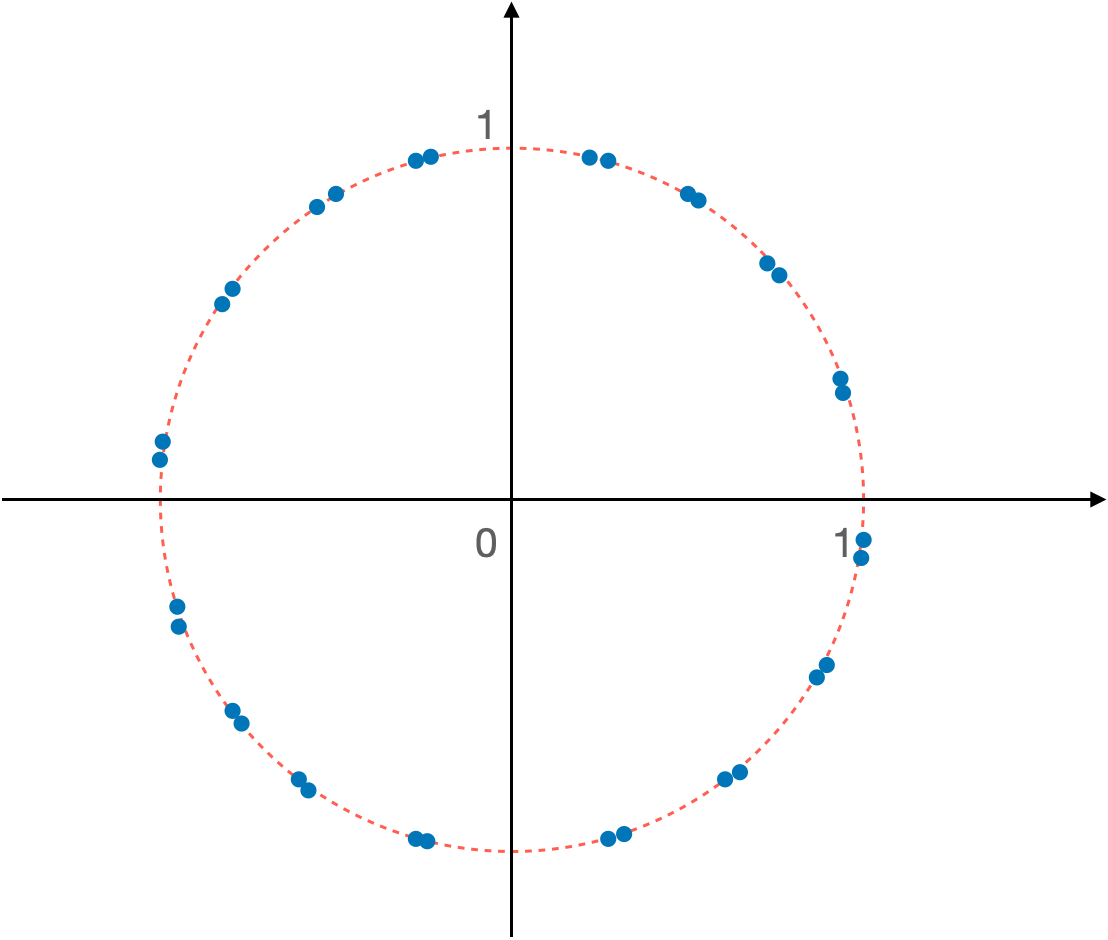}
\label{fig:uniformity}
\end{minipage}%
}%
\subfigure[dimensional collapse]{
\begin{minipage}[t]{0.22\linewidth}
\centering
\includegraphics[width=1\textwidth,angle=0]{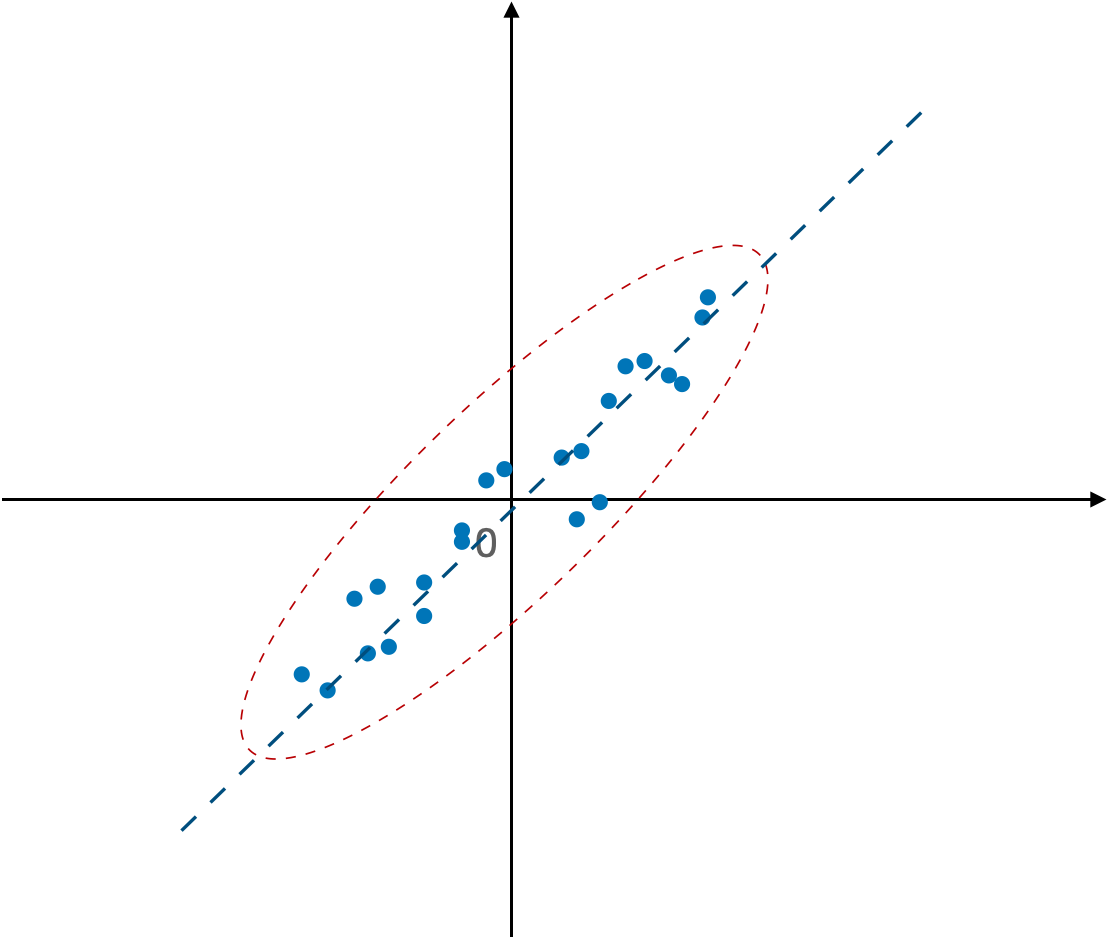}
\label{fig:dimensional}
\end{minipage}
}
\subfigure[decorrelation]{
\begin{minipage}[t]{0.22\linewidth}
\centering
\includegraphics[width=1\textwidth,angle=0]{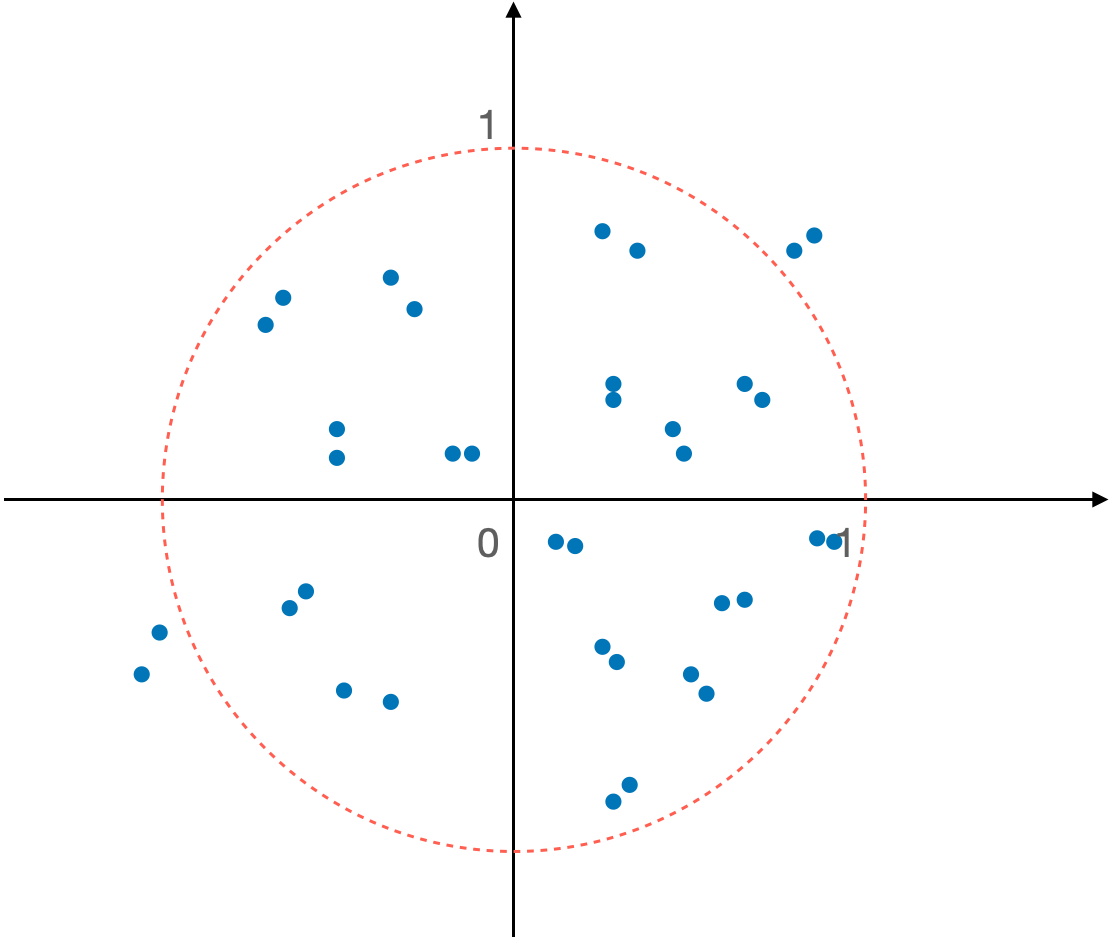}
\label{fig:decorrelation}
\end{minipage}
}
\caption{An illustration of the two types of collapse and how to deal with them with a 2-d case. Blue circles are data points. Fig.~\ref{fig:complete}: complete collapse, when all data samples degenerate to a same point on the hypersphere. Fig.~\ref{fig:uniformity}: the uniformity loss keeps positive pairs close, but forces all data points to distribute on the hypersphere uniformly. Fig.~\ref{fig:dimensional}: in dimensional collapse, the data points are not projected onto the hypersphere, but they distribute nearly as a line in the space, making them hard to discriminate. Fig.~\ref{fig:decorrelation}: the decorrelation term prevents dimensional collapse by directly decorrelating each dimensional representations, which implicitly scatters the data points.}
\label{fig:collapse}
\end{figure}

The uniformity term in Eq.~\eqref{eqn:uniform} prevents complete collapse by separating the embeddings of arbitrary two data points, so that the data points would be embedded uniformly on the hypersphere. Fig.~\ref{fig:complete} and \ref{fig:uniformity} provide an illustration for complete collapse and how the uniformity term prevents it.

Another kind of collapse that has been neglected by most existing works is \textbf{dimensional collapse}~\cite{decorrelation}. Different from complete collapse where all the data points degenerate into \textbf{a single point}, dimensional collapse means data points are distributed on \textbf{a line}, and each dimension captures exactly the same features (or different dimensions are highly correlated can capture the same information). Note that if the data representations are normalized along feature dimensions, all the data points would be projected onto a hypersphere. Under this circumstance there will not be dimensional collapse. However, if we normalize the output along the instance dimension so that each column has zero-mean and $1/\sqrt{N}$-standard deviation, as is done in this paper in Eq.~\eqref{eqn:loss-inv}, merely optimizing Eq.~\eqref{eqn:loss-inv} would not prevent dimensional collapse, i.e. $\tilde{\mathbf{Z}}^{\top}\tilde{\mathbf{Z}}  \cong \mathbf{1}$ ($\tilde{\mathbf{Z}} \in \mathbb{R}^{N \times D}$ is normalized by column).

In our model, the feature decorrelation term in Eq.~\eqref{eqn:loss-decor} exactly prevents dimensional collapse by minimizing $\left\| \tilde{\mathbf{Z}}^{\top} \tilde{\mathbf{Z}} - \mathbf{I} \right\|^2_F$. Note that the diagonal term is always equal to $1$, so we are pushing each dimension to capture orthogonal features. Also, the feature decorrelation term implicitly scatters the data points in the space, making them distinguishable for downstream tasks~\cite{decorrelation, whiten}. An illustration of the dimensional collapse and the effect of feature decorrelation is provided in Fig.~\ref{fig:dimensional} and \ref{fig:decorrelation}, respectively.

\section{Properties of Mutual Information and Entropy} \label{append:MI}
In this section, we enumerate some useful properties of mutual information and entropy that will be used in Appendix~\ref{append:proof} for proving the theorems. For any random variables $A, B, C, X, Z$, we have:
\begin{itemize}
    \item \textbf{Property 1}. Non-negativity:
    \begin{equation}
        I(A,B) \ge 0, I(A,B|C) \ge 0.
    \end{equation}
    
    \item \textbf{Property 2}. Chain rule:
    \begin{equation}
        I(A,B,C) = I(A,B) - I(A,B|C).
    \end{equation}
    
    \item \textbf{Property 3}. Data Processing Inequality (DPI). $Z = f_{\theta}(X)$, then:
    \begin{equation}
        I(Z, A) = I(f_{\theta}(X), A) \le I(X, A)
    \end{equation}
    
    \item \textbf{Property 4}. Non-negativity of discrete entropy. For discrete random variable:
    \begin{equation}
        H(A) \ge 0, H(A|B) \ge 0 .
    \end{equation}
    
    \item \textbf{Property 5}. Relationship between entropy and mutual information:
    \begin{equation}
        H(A) = H(A|B) + I(A,B).
    \end{equation}
    
    \item \textbf{Property 6}. Entropy of deterministic function. If $Z$ is deterministic given $X$:
    \begin{equation}
        H(Z|X) = 0
    \end{equation}
    
    \item \textbf{Property 7}. Entropy of Gaussian distribution. Assume $X$ obeys a $k$-dimensional Gaussian distribution, $X \sim \mathcal{N}({{\mu}, \Sigma})$, and we have
    \begin{equation}
        H(X) = \frac{k}{2}(\ln 2\pi + 1) + \frac{1}{2}\ln |\Sigma|.
    \end{equation}
    
\end{itemize}

\section{Proofs in Section~\ref{sec:insights}}\label{append:proof}
As already introduced in Section~\ref{sec:insights}, we use $X$ and $S$ to denote the data and its augmentations respectively. We use $Z_X$ and $Z_S$ to denote their embeddings through the encoder $f_{\theta}$: $Z_X = f_{\theta}(X)$, $Z_S = f_{\theta}(S)$. We aim to learn the optimal encoder parameters $\theta$.
\subsection{Proof of Proposition~\ref{prop:conditional-entropy}}
Restate Proposition~\ref{prop:conditional-entropy}:

\textbf{Proposition~\ref{prop:conditional-entropy}.} \textit{In expectation, minimizing Eq.~\eqref{eqn:loss-inv} is equivalent to minimizing the entropy of $Z_{S}$ conditioned on the input $X$, i.e.,:
    \begin{equation}
        \min\limits_{\theta} \mathcal{L}_{inv} \cong \min\limits_{\theta} H(Z_{S} | X)
    \end{equation}}
\begin{proof}
    Assume input data come from a distribution $\bm{x} \sim p(\bm{x})$ and $\bm{s}$ is a view of $\bm{x}$ through random augmentation $\bm{s} \sim p_{aug}(\cdot|\bm{x})$. Denote $\bm{z}_{\bm{s}}$ as the representation of $\bm{s}$. Note that $\bm{s}_1$ and and $\bm{s}_2$ both come from $p_{aug}(\cdot|\bm{x})$.

    Recall the invariance term: $\mathcal{L}_{inv} =  \left\| \tilde{\mathbf{Z}}_A - \tilde{\mathbf{Z}}_B \right\|_F^2 =  \sum\limits_{i=1}^{N}\sum\limits_{k=1}^D\left(\tilde{z}^A_{i,k}- \tilde{z}^B_{i,k}\right)^2 $. If we ignore the normalization and use $\bm{s}_1$ and $\bm{s}_2$ to represent view $A$ and view $B$. We have:
    \begin{equation}
    \begin{split}
        \mathcal{L}_{inv} / N \cong & \mathbb{E}_{\bm{x}}\left(\sum\limits_{k=1}^{D}\mathbb{E}_{\bm{s}_1, \bm{s}_2 \sim p(\cdot | \bm{x})}(z^{\bm{s}_1}_k - z^{\bm{s}_2}_k )^2 \right)\\
          = & \mathbb{E}_{\bm{x}} \left(  \sum\limits_{k=1}^{D} \mathbb{E}_{\bm{s}_1, \bm{s}_2 \sim p(\cdot | \bm{x})}({z^{\bm{s}_1}_k}^2 + {z^{\bm{s}_2}_k}^2 - 2* z^{\bm{s}_1}_k z^{\bm{s}_2}_k) \right)\\
          = & 2*\mathbb{E}_{\bm{x}}\left( \sum\limits_{k=1}^{D}  \mathbb{V}_{\bm{s}\sim p(\cdot|\bm{x})}z^{\bm{s}}_k \right) \\
          = & 2*\sum\limits_{k=1}^{D} \mathbb{E}_{\bm{x}} \left(\mathbb{V}_{\bm{s}\sim p(\cdot|\bm{x})}z^{\bm{s}}_k \right)
    \end{split}
    \end{equation}
This indicates that minimizing $\mathcal{L}_{inv}$ is to minimize the variance of augmentation's representations conditioned on the input data.

Note the decorrelation term Eq.~\eqref{eqn:loss-decor} aims to learn orthogonal representations at each dimension. If the representations are perfectly decorrelated, then $H(Z_S|X) = \sum\limits_{k}H(Z_{S,k}|X)$. With Assumption~\ref{assumption:gaussian}, each dimensional representation also obeys a 1-dimensional Gaussian distribution, whose entropy is $H(Z_{S,k}|X) = \frac{1}{2}\log 2 \pi e{\sigma_k^s}^2$. This indicates by minimizing the variance of features at each dimension, its entropy is also minimized. Hence we have Proposition~\ref{prop:conditional-entropy}.
\end{proof}

\begin{remark}\label{remark:entropy}
$I(Z_S, S|X) = H(Z_S|X) - H(Z_S|S,X) =  H(Z_S|X)$ (Property 6 in Appendix~\ref{append:MI}). So $I(Z_S,S|X)$ is also minimized.
\end{remark}

\subsection{Proof of Proposition~\ref{prop:entropy}}
Restate Proposition~\ref{prop:entropy}:

\textbf{Proposition~\ref{prop:entropy}.} \textit{Minimizing Eq.~\eqref{eqn:loss-decor} is equivalent to maximizing the entropy of $Z_S$, i.e.,
    \begin{equation}
        \min\limits_{\theta} \mathcal{L}_{dec} \cong \max\limits_{\theta} H(Z_{S}).
    \end{equation}}
\begin{proof}
    With the assumption that $Z_{S}$ obeys a Gaussian distribution, we have:
\begin{equation}
    \max\limits_{\theta} H(Z_{S}) \cong \max\limits_{\theta} \log |\Sigma_{Z_{S}}|,
\end{equation}
where $|\Sigma_{Z_{S}} |$ is the determinant of the covariance matrix of the embeddings of the augmented data. Note that in our implementation we normalize the embedding matrix along the instance dimension: $\Sigma_{Z_S} \cong \tilde{\mathbf{Z}}_S^{\top}\tilde{\mathbf{Z}}_S$, so the diagonal entries of $\Sigma_{Z_{S}}$ are all $1$'s. And $\Sigma_{Z_S} \in \mathbb{R}^{D\times D}$ is a symmetric matrix. 

If $\lambda_1, \lambda_2, \cdots, \lambda_D$ are the $D$ eigenvalues of $\Sigma_{Z_{S}}$, then $\sum\limits_{i=1}^D \lambda_i = \mathrm{trace}(\Sigma_{Z_{S}} )  = D$. We have
\begin{equation}
    \log |\Sigma_{Z_{\theta, X'}}| =  \log \prod\limits_{i=1}^{D} \lambda_i = \underbrace{\sum\limits_{i=1}^{D} \log \lambda_i \le D\log \frac{\sum\limits_{i=1}^D \lambda_i}{D}}_{\text{Jensen Inequality}}  = 0.
\end{equation}

This means that the upper bound of $|\Sigma_{Z_{S}}|$ is $1$, and the upper bound is achieved if and only if $\lambda_i = 1$ for $\forall i$, which indicates $\Sigma_{Z_{S}}$ is an identity matrix. This global optimum is exactly the same as that of the feature decorrelation term $\mathcal{L}_{dec}$ in Eq.~\eqref{eqn:loss-decor}. Therefore we conclude the proof.
\end{proof}
   
 \subsection{Proof of Theorem~\ref{the:mi}}
Restate Theorem~\ref{the:mi}:

\textbf{Theorem \ref{the:mi}}. \textit{ By optimizing Eqn~\eqref{eqn:loss}, we maximize the mutual information between the view's embedding $Z_S$ and the input data $X$, and minimize the mutual information between the view's embedding $Z_S$ and the view it self $S$, conditioned on the input data $X$. Formally,
   \begin{equation}
      \min\limits_{\theta} \mathcal{L} \Rightarrow \max\limits_{\theta} I(Z_S, X)\quad \; \mbox{and} \; \quad \min\limits_{\theta} I(Z_S, S|X).
   \end{equation}
 }
\begin{proof}
  According to Remark~\ref{remark:entropy}, we have:
  \begin{equation}
      I(Z_S,S|X) = H(Z_S|X).
  \end{equation}
  According to Property 5 in Appendix~\ref{append:MI}, we have:
  \begin{equation}
      I(Z_S, X) = H(Z_S) - H(Z_S|X).
  \end{equation}
Then combining Proposition~\ref{prop:conditional-entropy} and Proposition~\ref{prop:entropy}, we conclude the proof.
\end{proof}

\subsection{Proof of Theorem~\ref{the:ibssl}}\label{proof:ibssl}
Restate Theorem~\ref{the:ibssl}:

\textbf{Theorem 2.}
\textit{Assume $0 < \beta \le 1$, then by minimizing Eq.~\eqref{eqn:loss}, the self-supervised Information Bottleneck objective is maximized, formally:
   \begin{equation}
       \min\limits_{\theta} \mathcal{L} \Rightarrow \max\limits_{\theta} \mathcal{IB}_{ssl}.
   \end{equation}}
\begin{proof}
    According to Property 5 in Appendix~\ref{append:MI}, we can rewrite the IB principle in SSL setting as:
    \begin{equation}
            \mathcal{IB}_{ssl} = [H(Z_{S}) - H(Z_{S} | X)] - \beta\left[H(Z_{S}) - H(Z_{S}|S)\right].
    \end{equation}
    
    Notice that $Z_{S}$ is deterministic given $S$: $Z_S = f_{\theta}(S)$. According to Property 5 in Appendix~\ref{append:MI}, we have $H(Z_{S}|S) = 0$. Hence, we further have the following relationship
    \begin{equation}\label{eqn:ibssl}
            \mathcal{IB}_{ssl}  = (1-\beta) H(Z_{S}) - H(Z_{S}|X).
    \end{equation}
    
    Let $\lambda = 1-\beta \ge 0$. Now we can decompose the objective $\mathcal{IB}_{ssl}$ into two terms: 1) maximizing $H(Z_{S})$, which increases the information entropy of the embeddings of augmented data. 2) minimizing $H(Z_{S}|X)$, which decreases the entropy of the embeddings of augmented data, conditioned on the original data.
    
    With Proposition~\ref{prop:conditional-entropy} and Proposition~\ref{prop:entropy}, we complete the proof.
\end{proof}

\subsection{Proof of Corollary~\ref{corollary:1}}\label{proof:corollary-1}
Restate Corollary~\ref{corollary:1}:

\textbf{Corollary~\ref{corollary:1}.}
\textit{Let $X_1 = S$, $X_2 = X$ and assume $0< \beta \le 1$, then minimizing Eq.~\eqref{eqn:loss} is equivalent to minimizing the Multi-view Information Bottleneck Loss in~\cite{Multi-IB}:
    \begin{equation}
       \mathcal{L}_{MIB} = I(Z_1, X_1 | X_2) - \beta I(X_2, Z_1), 0 <\beta \le 1
    \end{equation}}
    
By maximizing $I(X_2, Z_1)$, the model could obtain sufficient information for downstream tasks by ensuring the representation $Z_1$ of $X_1$ is sufficient for $X_2$, and decreasing $I(Z_1, X_1 |X_2)$ will increase the robustness of the representation by discarding irrelevant information.
\begin{proof}
Let $X_1 = S, X_2 = X$ be two views of the input data. We have:
\begin{equation}
\begin{split}
    \mathcal{L}_{MIB} & = I(Z_S, S|X) - \beta I(X, Z_S) \\
                      & = [H(Z_S|X) - H(Z_S|S,X)] - \beta[H(Z_S) - H(Z_S|X)] \\
                      & = (1-\beta)H(Z_S|X) -\beta{H(Z_S)} - H(Z_S|S,X). \\
\end{split}
\end{equation}
As $Z_S$ is deterministic given $S$, we can obtain $H(Z_S|S, X) = 0$. Based on this, we can further simplify $\mathcal{L}_{MIB}$ as
\begin{equation}\label{eqn:ibmib}
    \mathcal{L}_{MIB} = H(Z_S|X) - \lambda H(Z_S), \;\mbox{with}\;\lambda > 0.
\end{equation}
With Proposition~\ref{prop:conditional-entropy} and Proposition~\ref{prop:entropy}, we complete the proof.
\end{proof}

\subsection{Proof of Corollary~\ref{corollary:2}}\label{proof:corollary-2}
Restate Corollary~\ref{corollary:2}:

\textbf{Corollary~\ref{corollary:2}.}
\textit{ When the data augmentation process is reversible, minimizing Eq.~\eqref{eqn:loss} is equivalent to learning the Minimal and Sufficient Representations for Self-supervision in~\cite{mini-1}:
     \begin{equation}\label{eqn:ibmini}
        Z^{\text{ssl}}_{X} = \mathop{\arg \max}\limits_{Z_X} I(Z_X, S), Z^{\text{ssl}_{\text{min}}}_X = \mathop{\arg \min}\limits_{Z_X} H(Z_X | S)  \; \mbox{s.t.} \; I(Z_X, S) \text{ is maximized}.
    \end{equation}}
    
$Z_{X}^{ssl}$ is the sufficient self-supervised representation by maximizing $I(Z_X, S)$, and $Z_{X}^{\text{ssl}_{\text{min}}}$ is the minimal and sufficient representation by minimizing $H(Z_X|S)$.
\begin{proof}
Eq.~\eqref{eqn:ibmini} can be converted to minimizing the relaxed Lagrangian objective as below
\begin{equation}
    \mathcal{L}_{{\text{ssl}}_{\text{min}}} = H(Z_X|S) - \beta I(Z_X, S),\;\mbox{with}\; \beta > 0.
\end{equation}
Then $\mathcal{L}_{\text{ssl}_{min}}$ could be decomposed into
\begin{equation}
\begin{split}
    \mathcal{L}_{{\text{ssl}}_{\text{min}}} & = H(Z_X|S) - \beta I(S, Z_X) \\
                      & = H(Z_X|S) - \beta[H(Z_X) - H(Z_X|S)] \\
                      & = (1+\beta)H(Z_X|S) -\beta{H(Z_X)} \\
\end{split}
\end{equation}
With $\beta > 0$, $\mathcal{L}_{\text{ssl}_{min}}$ is essentially a symmetric formulation of Eq.~\eqref{eqn:ibssl}, by exchanging $X$ with $S$, and $Z_X$ with $Z_S$. With the assumption that the data augmentation process is reversible and according to Proposition~\ref{prop:conditional-entropy} and Proposition~\ref{prop:entropy}, we conclude the proof.
\end{proof}
\subsection{Proof of Theorem~\ref{the:downstream}}
\label{proof:downstream}
Restate Theorem~\ref{the:downstream}:

\textbf{Theorem~\ref{the:downstream}}
\textit{(task-relevant/irrelevant information). By optimizing Eq.~\eqref{eqn:loss}, the task-relevant information $I(Z_S, T)$ is maximized, and the task-irrelevant information $H(Z_S|T)$ is minimized. Formally:
  \begin{equation}
      \min\limits_{\theta} \mathcal{L} \Rightarrow \max\limits_{\theta} I(Z_S,T) \;\mbox{and}\;\min\limits_{\theta} H(Z_S | T).
  \end{equation}}
\begin{proof}
Note that with Assumption~\ref{assumption:redundancy}, we have $I(X,T|S) = I(S,T|X) = 0$. Therefore we obtain $0 \le I(Z_S,T|X) \le I(S, T|X) = 0$, which induces $I(Z_S,T|X) = 0$. Then we can derive
\begin{equation}\label{eqn:task-relevant}
\begin{split}
    I(Z_S, T) = & I(Z_S,T|X) + I(Z_S,X,T) \\
              = & 0 + I(Z_S, X) - I(Z_S, X | T) \\
              = & I(Z_S,X) - I(Z_S, X|T) \\
              \ge & I(Z_S,X) - I(X,S|T) \\
\end{split}
\end{equation}
and 
\begin{equation}\label{eqn:task-irrelevant}
\begin{split}
    H(Z_S|T) =  & H(Z_S|X,T) + I(Z_S,X|T) \\
             =  & H(Z_S|X) - I(Z_S,T|X) + I(Z_S,X|T)  \\
             =  & H(Z_S|X)- 0 + I(Z_S,X|T) \\
              \le & H(Z_S|X) + I(X,S|T) \\
\end{split}
\end{equation}
Note that $I(X,S|T)$ is a fixed gap indicating the amount of task-irrelevant information shared between $X$ and $S$.

With Theorem~\ref{the:mi}, by optimizing the objective Eq.~\eqref{eqn:loss}, we maximize the lower bound of the task-relevant information $I(Z_S, T)$, and minimize the upper bound of the task-irrelevant information $H(Z_S|T)$. Then the proof is completed.
\end{proof}
\section{Implementation Details}\label{append:exp}
\subsection{Loss function}
In our implementation we did not directly use the original loss function as given in Eqn.~\eqref{eqn:loss}. For simplicity, we use its equivalent form, which can be easily derived from the following equation:
\begin{equation}\label{eqn:implementation-loss-derivation}
\begin{split}
    \left\|\tilde{\mathbf{Z}}_A  - \tilde{\mathbf{Z}}_B \right\|^2_F = &  \sum\limits_{k=1}^{D}\sum\limits_{i=1}^{N} (\tilde{\bm{z}}^A_{i,k} - \tilde{\bm{z}}^B_{i,k})^2  \\
     = & \sum\limits_{k=1}^{D}\sum\limits_{i=1}^{N} \left(({\tilde{\bm{z}}^A_{i,k}})^2 + ({\tilde{\bm{z}}^B_{i,k}})^2  - 2 *  {\tilde{\bm{z}}^A_{i,k}}{\tilde{\bm{z}}^B_{i,k}}\right)    \\
     = & \sum\limits_{k=1}^{D}(2 - 2* \tilde{Z}_{A,k}^{\top}\tilde{Z}_{B,k} ) \\
     = & 2D - 2*\mathrm{trace}(\tilde{\bm{Z}}_{A}^{\top}\tilde{\bm{Z}}_{B})
     \end{split}
\end{equation}

So we can rewrite the objective function Eqn.~\eqref{eqn:loss} as the following one:
\begin{equation}\label{eqn:implementation-loss}
    \mathcal{L} = - \mathrm{trace}(\tilde{\bm{Z}}_{A}^{\top}\tilde{\bm{Z}}_{B}) + \lambda' \left( \left\|\tilde{\mathbf{Z}}_A^{\top}\tilde{\mathbf{Z}}_A - \mathbf{I}\right\|_F^2 + \left\|\tilde{\mathbf{Z}}_B^{\top}\tilde{\mathbf{Z}}_B - \mathbf{I}\right\|_F^2 \right)
\end{equation}
where the $\lambda'$ here should be half of the $\lambda$ in Eqn.~\eqref{eqn:loss}. For simplicity we do not discriminate between these two symbols. The values of the trade-off parameter $\lambda$ in Fig~\ref{fig:abl-lam} as well as that in Appendix~\ref{append:hyperparamters} are actually denoted as $\lambda'$ in Eqn.~\eqref{eqn:implementation-loss}.

\begin{table}[tb!]
	\centering
	\caption{Statistics of benchmark datasets}
	\label{tbl-statistics}
	\small
	\begin{threeparttable}
	\setlength{\tabcolsep}{4mm}{
		\begin{tabular}{ccccc}
			\toprule[0.8pt]
			Dataset  & \#Nodes & \#Edges & \#Classes & \#Features \\
			\midrule[0.6pt]
            Cora     & 2,708   &  10,556 &  7 & 1,433\\
            Citeseer & 3,327   &  9,228  & 6  & 3,703 \\
            Pubmed   & 19,717  &  88,651 & 3  & 500 \\
            Coauthor CS       & 18,333  &  327,576  & 15  & 6,805 \\
            Coauthor Physics  & 34,493  &  991,848  & 5  & 8,451 \\
			Amazon Computer & 13,752 & 574,418 & 10 & 767\\
			Amazon Photo   & 7,650 & 287,326 & 8  & 745 \\
			\bottomrule[0.8pt]
		\end{tabular}}
	\end{threeparttable}
\end{table}

\subsection{Graph augmentations}
We adopt two random data augmentations strategies on graphs: 1) \textbf{Edge dropping}. 2) \textbf{Node feature masking}. The two strategies are widely used in node-level contrastive learning~\cite{grace, grace-ad, bgrl}.

\begin{itemize}
    \item \textbf{Edge dropping} works on the graph structure level, where we randomly remove a portion of edges in the original graph. Formally, given the edge dropping ratio $p_e$, for each edge we have $p_e$ probability to drop this edge from the graph. When calculating the degree for each node, the dropped edge will not be considered.
    \item \textbf{Node feature masking} works on the node feature level, where we randomly set a fraction of features of all nodes as $0$. Formally, given the node feature masking ratio $p_f$, for each input feature, we set it as $0$ with a probability of $p_f$. Note that the masking operation is applied to the selected feature columns of all the nodes.
\end{itemize}

Note that the previous works~\cite{grace,grace-ad,bgrl} use two separate sets of edge dropping ratio $p_e$ and node feature dropping ratio $p_f$ for generating two views, i.e. $p_{e_1}$ and $p_{f_1}$ for view $A$,  $p_{e_2}$ and $p_{f_2}$ for view $B$. However, in our implementation, we let $p_{e_1} = p_{e_2}$ and $p_{f_1} = p_{f_2}$, so that the two transformations $t_A$ and $t_B$ come from the same distribution $\mathcal{T}$.

\subsection{Datasets}\label{append:exp-dataset}
We evaluate our models on seven node classification benchmarks: \textit{Cora, Citeseer, Pubmed, Coauthor CS, Coauthor Physics, Amazon Computer} and \textit{Amazon Photo}. We provide dataset statistics in Table~\ref{tbl-statistics}, and brief introduction and settings are as follows:

\textbf{Cora\footnote{ \url{https://relational.fit.cvut.cz/dataset/CORA}}, Citeseer, Pubmed}\footnote{Citeseer and Pubmed:  \url{https://linqs.soe.ucsc.edu/data}} are three widely used node classification benchmarks~\cite{dataset-cora-citeseer,dataset-pubmed}. Each dataset contains one citation network, where nodes mean papers and edges mean citation relationships. We use the public split for linear evaluation, where each class has fixed $20$ nodes for training, another fixed $500$ nodes and $1000$ nodes are for validation/test respectively.

\textbf{Coauther CS, Coauther Physics} are co-authorship graphs based on the Microsoft Academic Graph from the KDD Cup 2016 challenge \cite{dataset-coauther}. Nodes are authors, that are connected by an edge if they co-authored a paper; node features represent paper keywords for each author’s papers, and class labels indicate most active fields of study for each author. As there is no public split for these datasets, we randomly split the nodes into train/validation/test (10\%/10\%/80\%) sets.

\textbf{Amazon Computer, Amazon Photo} are segments of the Amazon co-purchase graph \cite{dataset-amazon}, where nodes represent goods, edges indicate that two goods are frequently bought together; node features are bag-of-words encoded product reviews, and class labels are given by the product category. We also use a 10\%/10\%/80\% split for these two datasets.

For all datasets, we use the processed version provided by Deep Graph Library~\cite{dgl}\footnote{\url{https://docs.dgl.ai/en/0.6.x/api/python/dgl.data.html}, Apache License 2.0}. All datasets are public available and do not have licenses.

In Table~\ref{tbl-exp-citation}, we have mentioned that for MVGRL~\cite{mvgrl} and GRACE~\cite{grace}, we reproduce the experiments with authors' codes, both of which are publicly available: MVGRL\footnote{\url{https://github.com/kavehhassani/mvgrl}, no license.} and GRACE\footnote{\url{https://github.com/CRIPAC-DIG/GRACE}, Apache License 2.0.}. 

\subsection{Hyper-parameters}\label{append:hyperparamters}
We provide all the detailed hyper-parameters on the seven benchmarks in Table~\ref{tbl-para}. All hyper-parameters are selected through small grid search, and the search space is provided as follows:
\begin{itemize}
    \item Training steps: \{20, 50, 100, 200\}
    \item Number of layers: \{1, 2, 3\}
    \item Number of hidden units: \{128, 256, 512, 1024\}
    \item $\lambda$: \{1e-4, 5e-4, 1e-3, 5e-3, 1e-2\}
    \item learning rate of CCA-SSG: \{5e-4, 1e-3, 5e-3\}
    \item weight decay of CCA-SSG: \{0, 1e-5, 1e-4, 1e-3\}
    \item edge dropping ratio: \{0, 0.1, 0.2, 0.3, 0.4, 0.5\}
    \item node feature masking ratio: \{0, 0.1, 0.2, 0.3, 0.4, 0.5\}
    \item learning rate of logistic regression: \{1e-3, 5e-3, 1e-2\}
    \item weight decay of logistic regression: \{1e-4, 1e-3, 1e-2\}
\end{itemize}


\begin{table}[t]
	\centering
	\caption{Details of hyper-parameters of the experimental results in Table~\ref{tbl-exp-citation} and Table~\ref{tbl-exp-co}.}
	\label{tbl-para}
	\small
	\begin{threeparttable}
{ 
		\begin{tabular}{ccccccccc|cc}
		    \toprule[1pt]
			 \multirow{2}{*}{Dataset} & \multicolumn{8}{c|}{CCA-SSG} &  \multicolumn{2}{c}{Logistic Regression} \\
			 \cline{2-9} \cline{10-11}
			 ~    & Steps & \# layers &\# hidden units & $\lambda$ & lr & wd & $p_f$ & $p_e$  & lr   & wd  \\
			 \hline
			 Cora  & 50 & 2 & 512-512 & 1e-3 & 1e-3 & 0 & 0.1 & 0.4 & 1e-2 & 1e-4  \\
			 Citeseer  & 20 &  1 & 512 & 5e-4 & 1e-3 & 0 & 0.0 & 0.4 &   1e-2 & 1e-2  \\
			 Pubmed & 100 &  2 & 512-512 & 1e-3 & 1e-3 & 0 & 0.3 & 0.5 &  1e-2 & 1e-4   \\
			 Computer & 50 &  2 & 512-512 & 5e-4 & 1e-3 & 0 & 0.1 & 0.3 &  1e-2 & 1e-4  \\
			 Photo & 50 &  2 & 512-512 & 1e-3 & 1e-3 & 0 &  0.2 & 0.3 &  1e-2 & 1e-4   \\
			 CS$^{1}$ & 50 &  2 & 512-512 & 1e-3 & 1e-3 & 0 & 0.2 & - &  5e-3 & 1e-4  \\
			 Physics & 100 &  2 & 512-512 & 1e-3 & 1e-3 & 0 & 0.5 & 0.5 &  5e-3 & 1e-4   \\
			\bottomrule[1pt]
		\end{tabular}}
		\begin{tablenotes}
		\item[1] We use MLP (instead of GCN) as the encoder on Coauthor-CS, which is essentially equivalent to setting $p_e = 1.0$ (drop all the edges except the self-loops). 
        \end{tablenotes}
	\end{threeparttable}
\end{table}

\section{Additional Experiments}\label{append:add}
\subsection{Visualizations of Correlation Matrix}
\begin{figure}[tb!]
\centering
\subfigure[Cora: raw feature]{
\begin{minipage}[t]{0.3\linewidth}
\centering
\includegraphics[width=0.9\textwidth,angle=0]{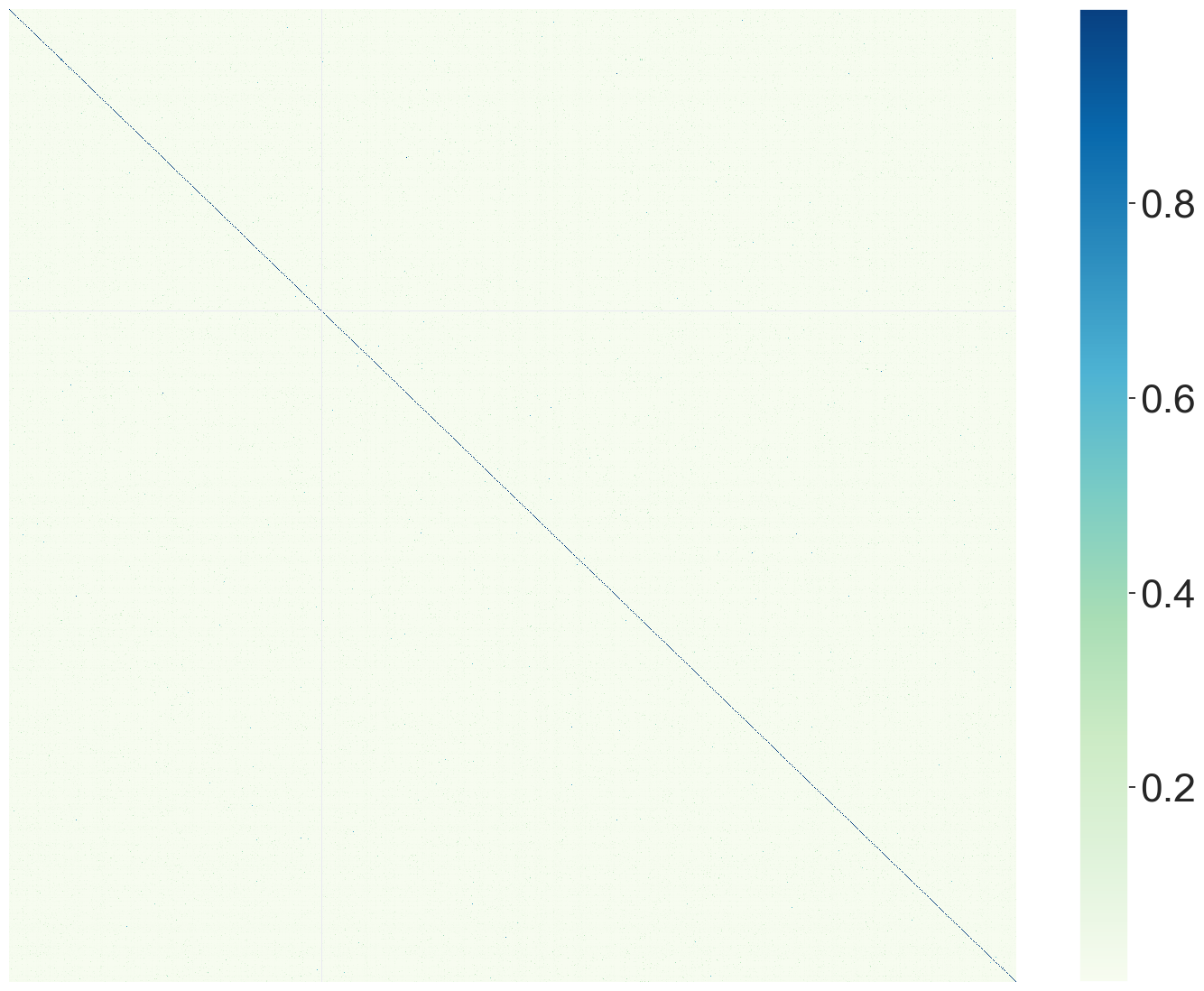}
\label{vis:attn-cora-1}
\end{minipage}%
}%
\subfigure[Cora: w/o decorrelation]{
\begin{minipage}[t]{0.3\linewidth}
\centering
\includegraphics[width=0.9\textwidth,angle=0]{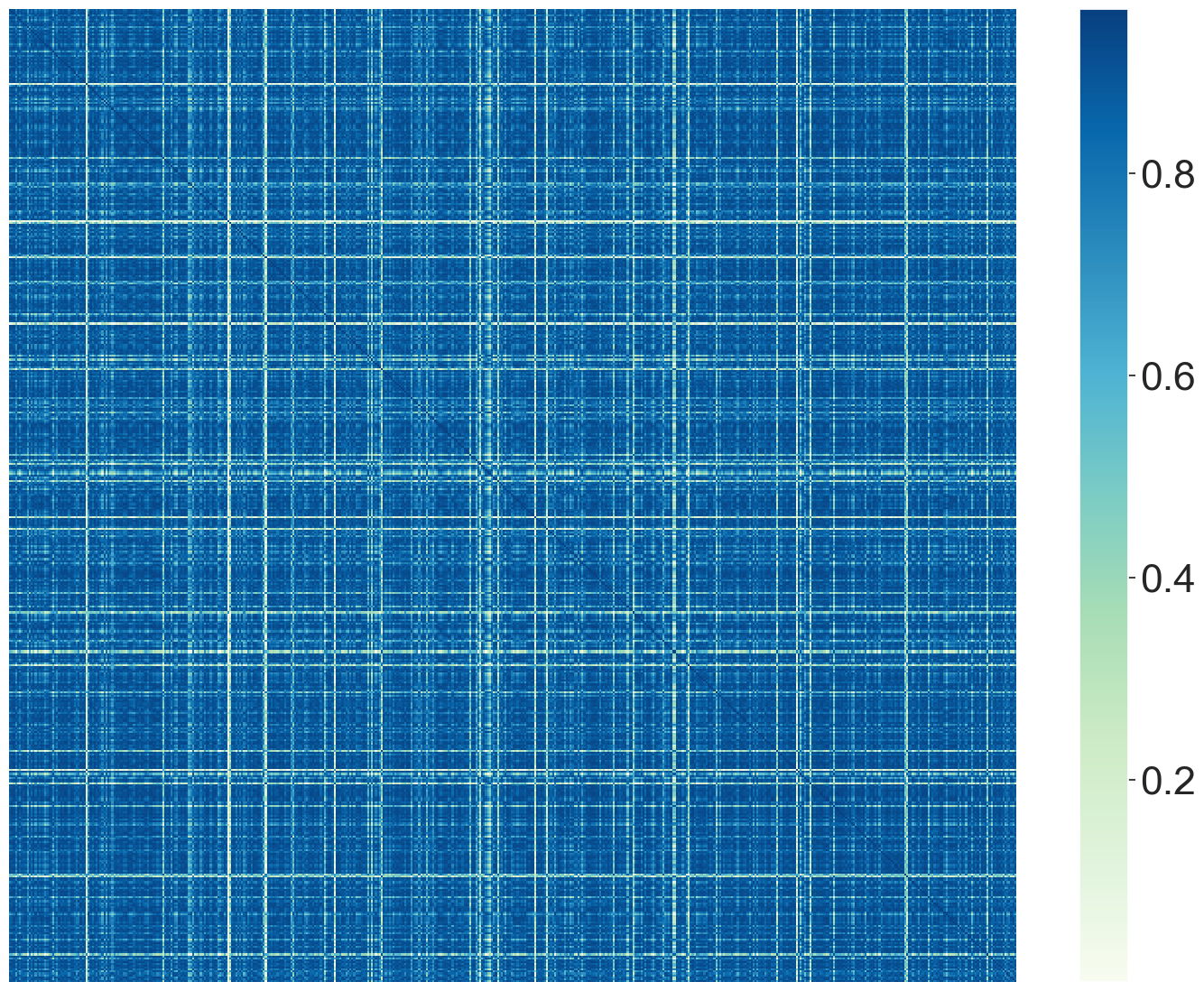}
\label{vis:attn-cora-2}
\end{minipage}%
}%
\subfigure[Cora: with decorrelation]{
\begin{minipage}[t]{0.3\linewidth}
\centering
\includegraphics[width=0.9\textwidth,angle=0]{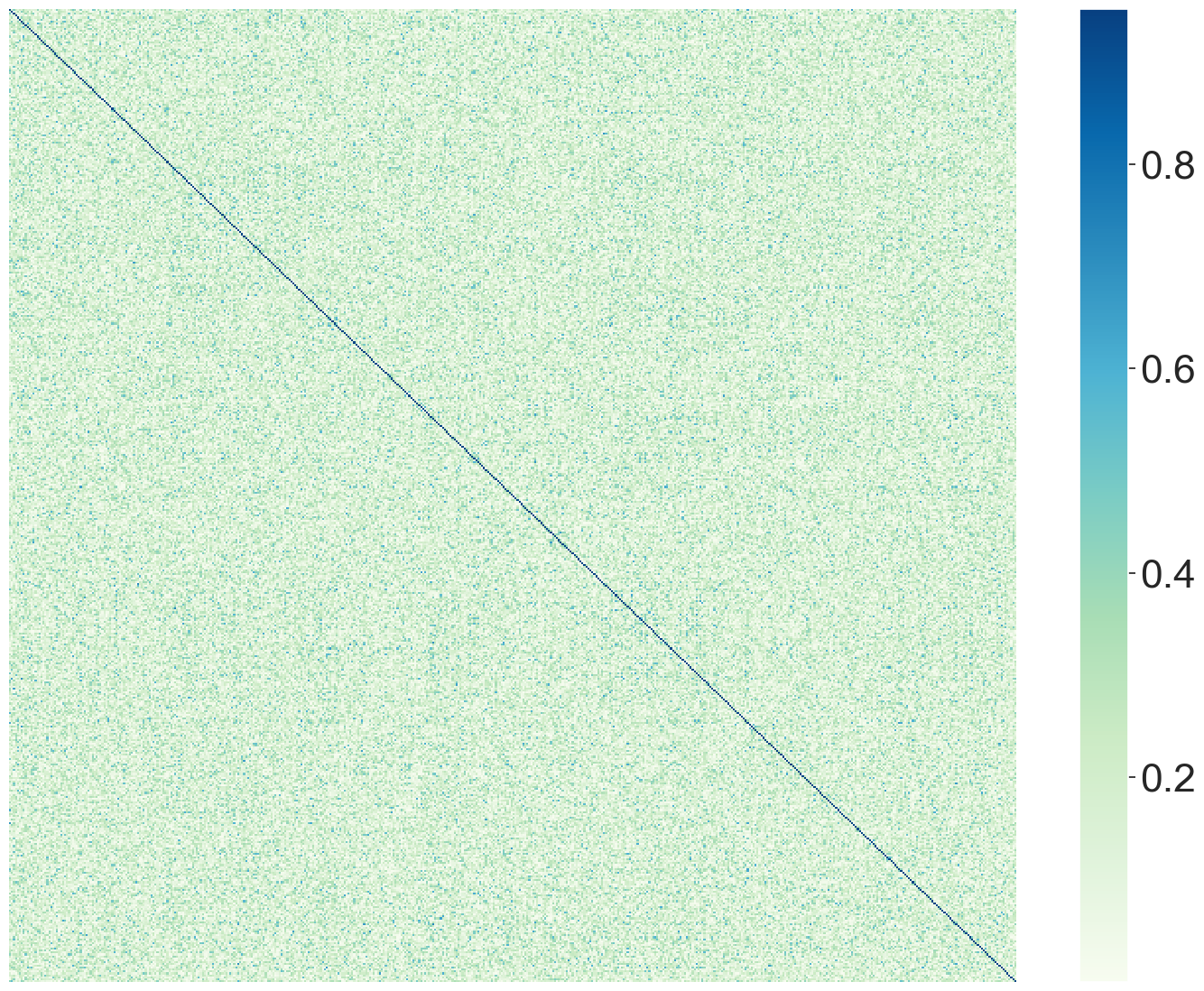}
\label{vis:attn-cora-3}
\end{minipage}
}

\subfigure[Citeseer: raw feature]{
\begin{minipage}[t]{0.3\linewidth}
\centering
\includegraphics[width=0.9\textwidth,angle=0]{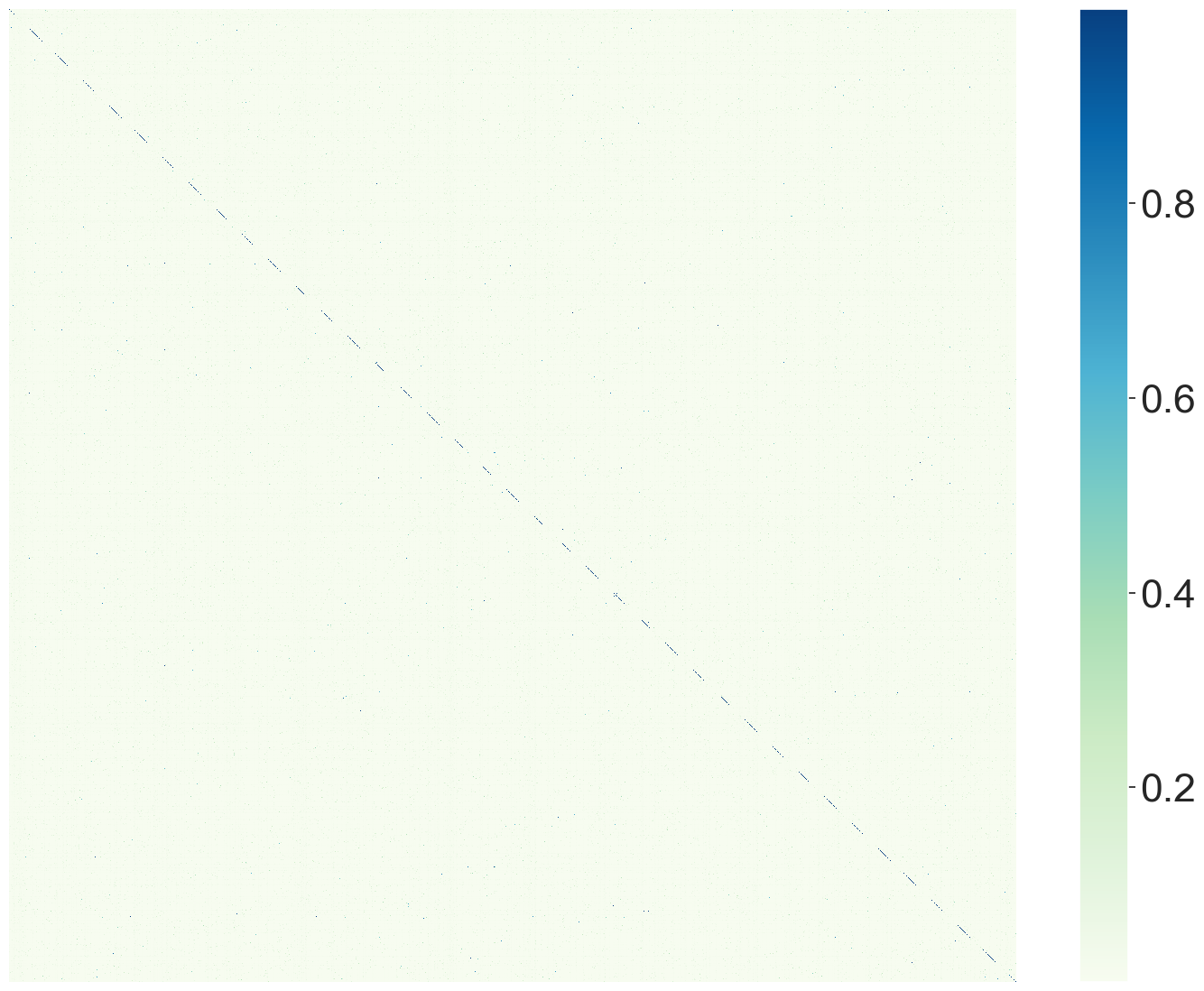}
\label{vis:attn-citeseer-1}
\end{minipage}%
}%
\subfigure[Citeseer: w/o decorrelation]{
\begin{minipage}[t]{0.3\linewidth}
\centering
\includegraphics[width=0.9\textwidth,angle=0]{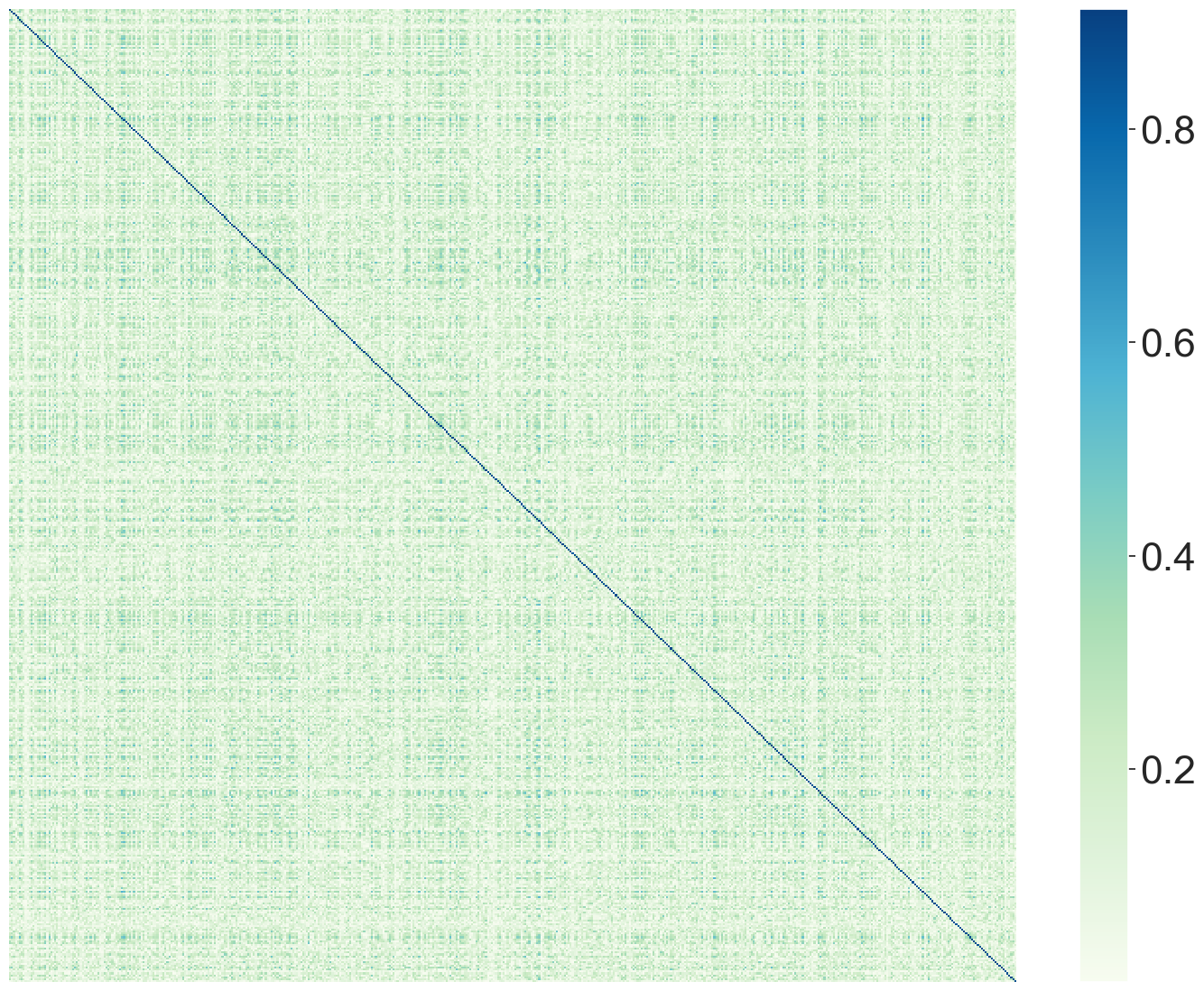}
\label{vis:attn-citeseer-2}
\end{minipage}%
}%
\subfigure[Citeseer: with decorrelation]{
\begin{minipage}[t]{0.3\linewidth}
\centering
\includegraphics[width=0.9\textwidth,angle=0]{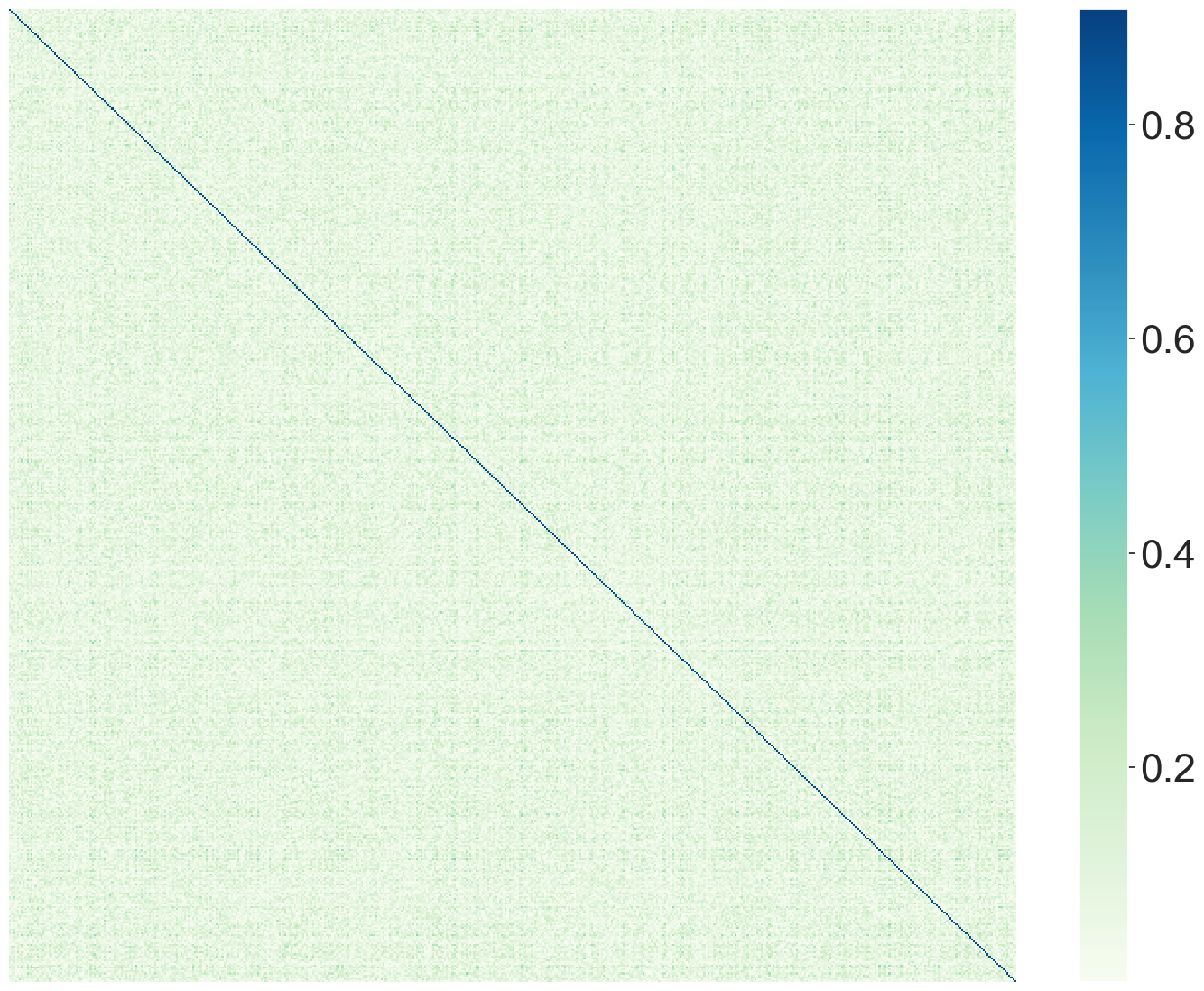}
\label{vis:attn-citeseer-3}
\end{minipage}
}

\subfigure[Pubemd: raw feature]{
\begin{minipage}[t]{0.3\linewidth}
\centering
\includegraphics[width=0.9\textwidth,angle=0]{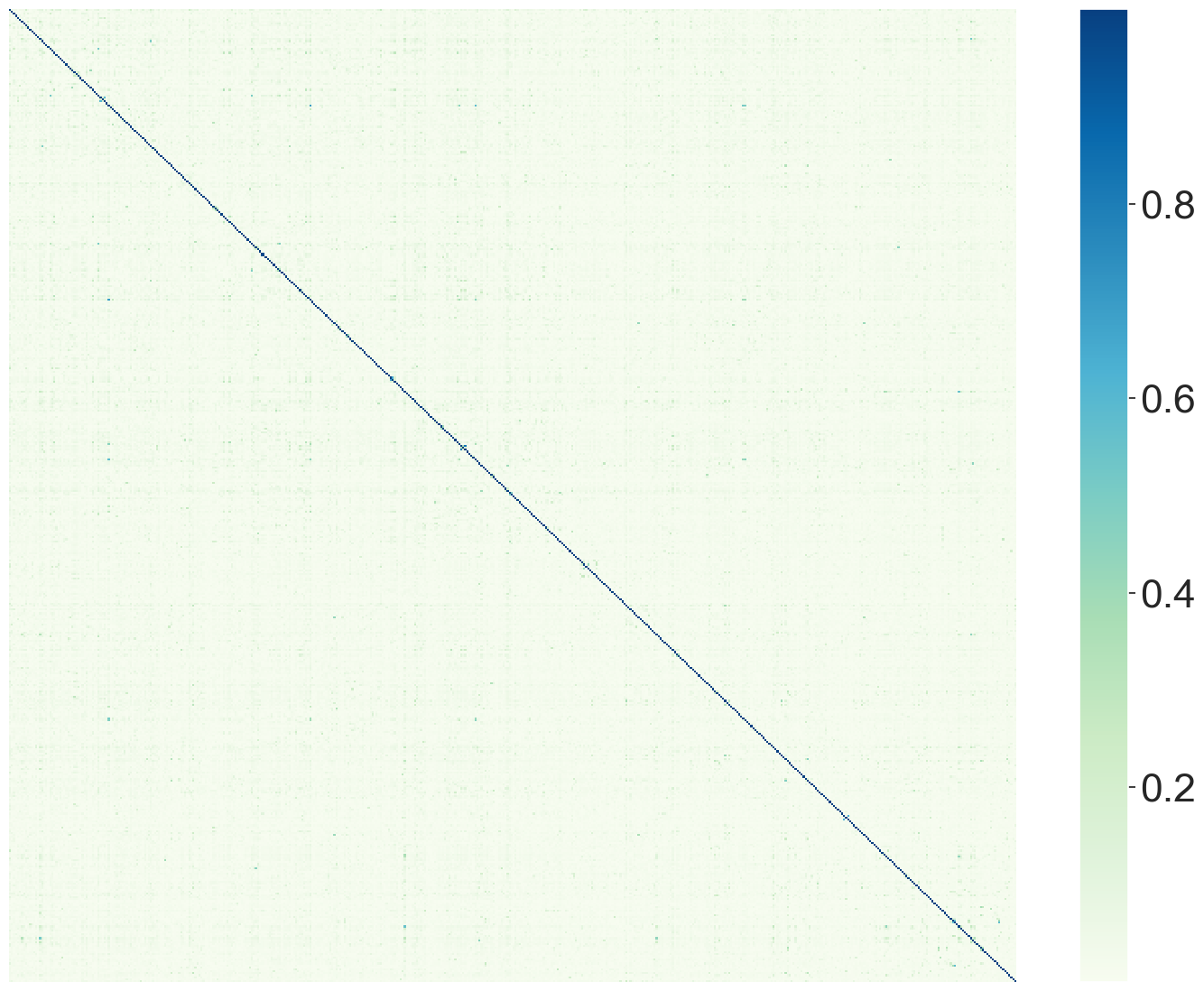}
\label{vis:attn-pubmed-1}
\end{minipage}%
}%
\subfigure[Pubmed: w/o decorrelation]{
\begin{minipage}[t]{0.3\linewidth}
\centering
\includegraphics[width=0.9\textwidth,angle=0]{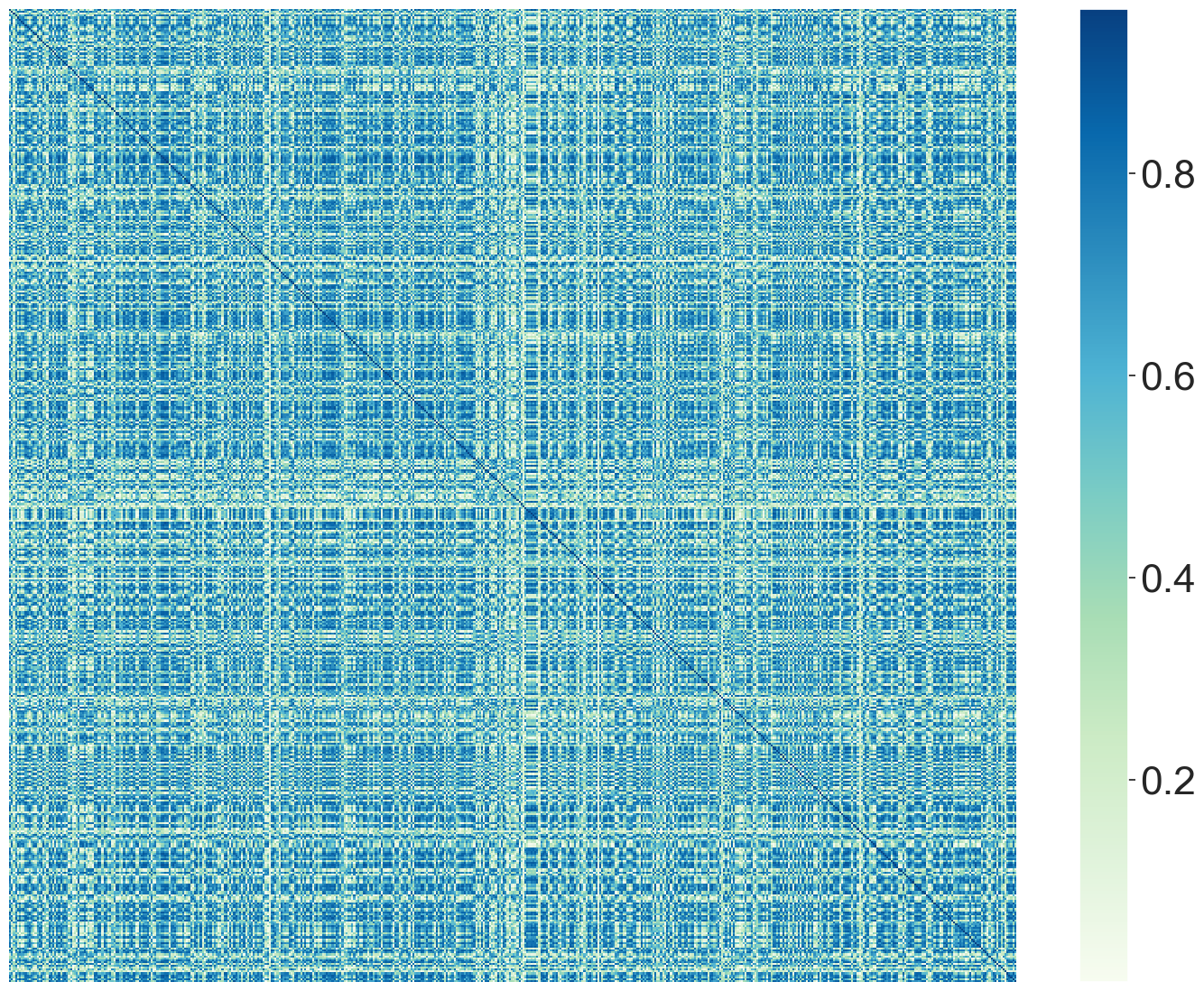}
\label{vis:attn-pubmed-2}
\end{minipage}%
}%
\subfigure[Pubmed: with decorrelation]{
\begin{minipage}[t]{0.3\linewidth}
\centering
\includegraphics[width=0.9\textwidth,angle=0]{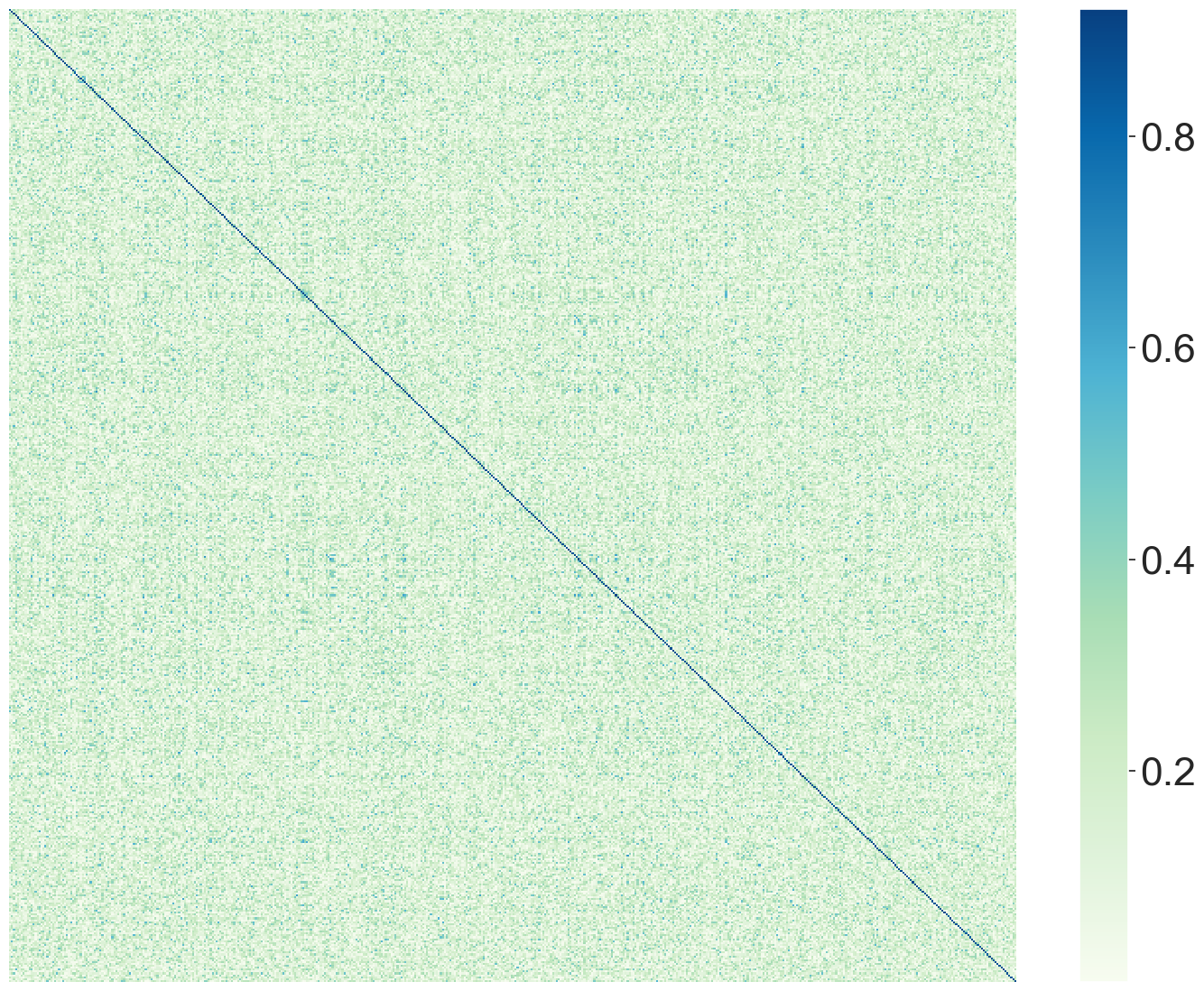}
\label{vis:attn-pubmed-3}
\end{minipage}
}
\caption{Visualizations of the correlation matrix (absolute value) of the raw input features, the embeddings without decorrelation term, and embeddings with decorrelation term on \textit{Cora, Citeseer} and \textit{Pubmed}. Light green: $\rightarrow 0$; Dark blue: $\rightarrow 1$.}
\label{fig:vis-attn}
\end{figure}

In Fig.~\ref{fig:vis-attn}, we provide visualizations of the absolute correlation matrix of the raw input features, the embeddings without decorrelation term, and embeddings with decorrelation term on three datasets: \textit{Cora, Citeseer} and \textit{Pubmed}.

As we can see, the raw input feature of the three datasets are all nearly fully uncorrelated (Fig.~\ref{vis:attn-cora-1}, \ref{vis:attn-citeseer-1} and \ref{vis:attn-pubmed-1}). Specifically, the on-diagonal term is close to $1$ while the off-diagonal term is close to $0$. When training without the decorrelation term~Eq.~\eqref{eqn:loss-decor}, the off-diagonal elements of the correlation matrix of node embeddings increase dramatically as shown in Fig.~\ref{vis:attn-cora-2} and \ref{vis:attn-pubmed-2}, indicating that different dimensions fail to capture orthogonal information. Fig.~\ref{vis:attn-cora-3} and \ref{vis:attn-pubmed-3} show that with the decorrelation term Eq.~\eqref{eqn:loss-decor}, our method could learn nearly highly disentangled representations. An interesting finding is that even without the decorrelation term, on Citeseer our method could still generate fairly uncorrelated representations (Fig.~\ref{vis:attn-citeseer-2}). The possible reason is that: 1) on Citeseer, we use a one-layer GCN as the encoder, which is less expressive than a two-layer one and alleviate the trend of collapsing. 2) The number of training steps on Citeseer is much smaller than others, so that the impact of invariance term is weaken.

These visualizations also echo the dimensional collapse issue as discussed in Appendix~\ref{append:collapse}: without the feature decorrelation term Eq.~\eqref{eqn:loss-decor}, there is a high probability that all the dimensions capture similar semantic information, thus leading to the dimensional collapse issue. The dimensional collapse can be fundamentally avoided by the decorrelation term Eq.~\eqref{eqn:loss-decor}.

\subsection{Effects of Augmentation Intensity}
\begin{figure}[tb!]
\centering
\subfigure[Cora]{
\begin{minipage}[t]{0.3\linewidth}
\centering
\includegraphics[width=0.77\textwidth,angle=0]{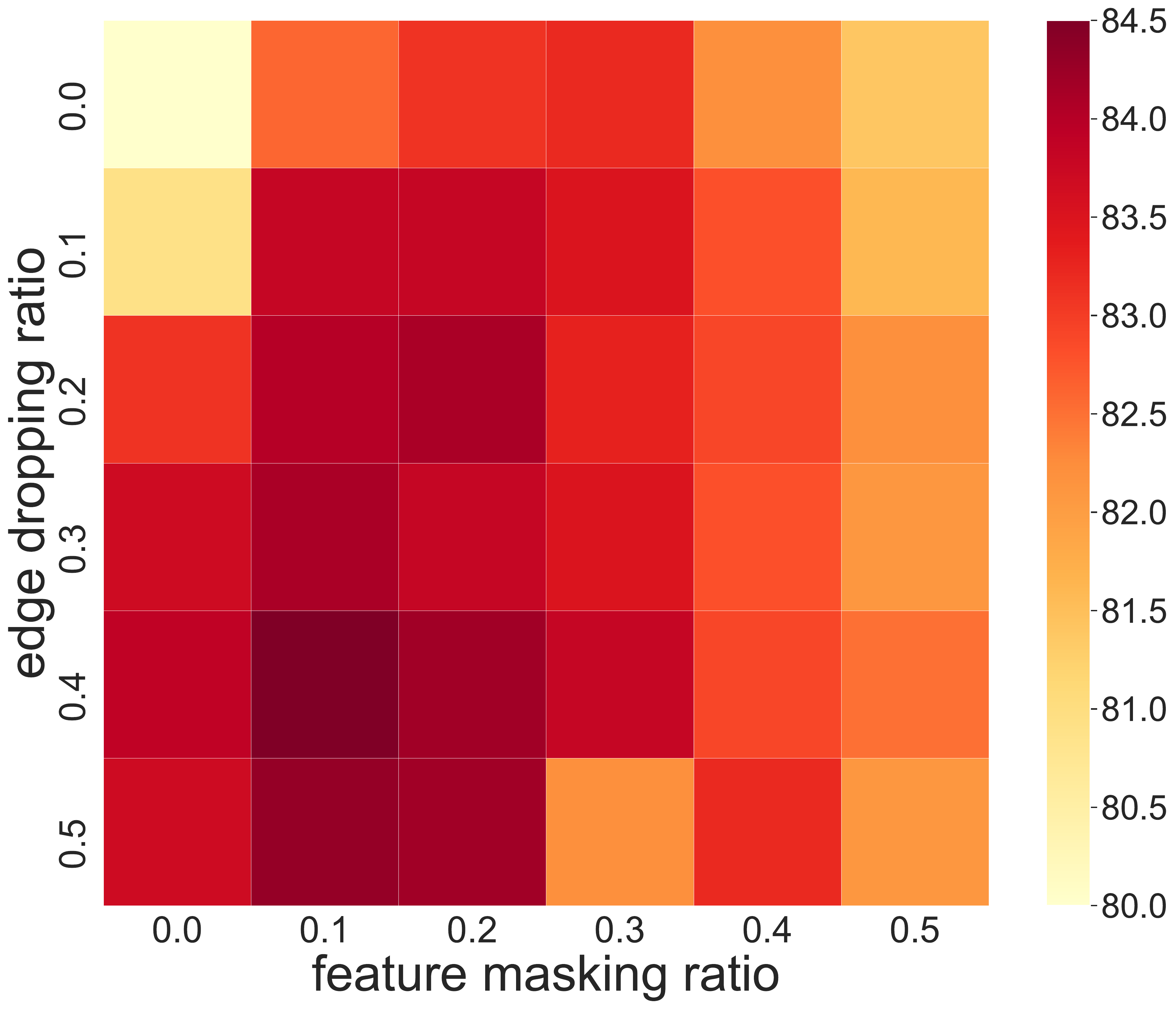}
\end{minipage}%
}%
\subfigure[Citeseer]{
\begin{minipage}[t]{0.3\linewidth}
\centering
\includegraphics[width=0.77\textwidth,angle=0]{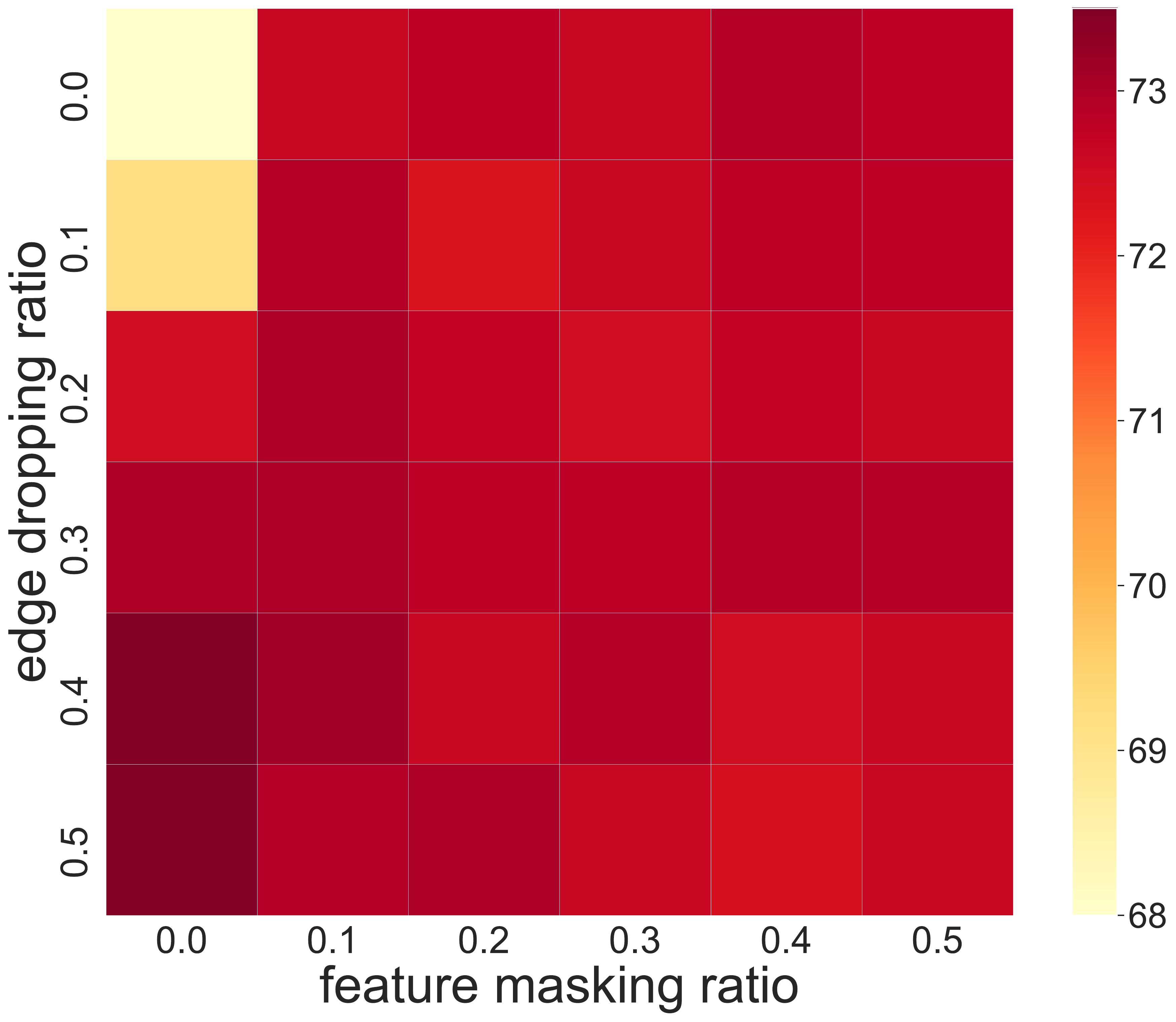}
\end{minipage}%
}%
\subfigure[Pubmed]{
\begin{minipage}[t]{0.3\linewidth}
\centering
\includegraphics[width=0.77\textwidth,angle=0]{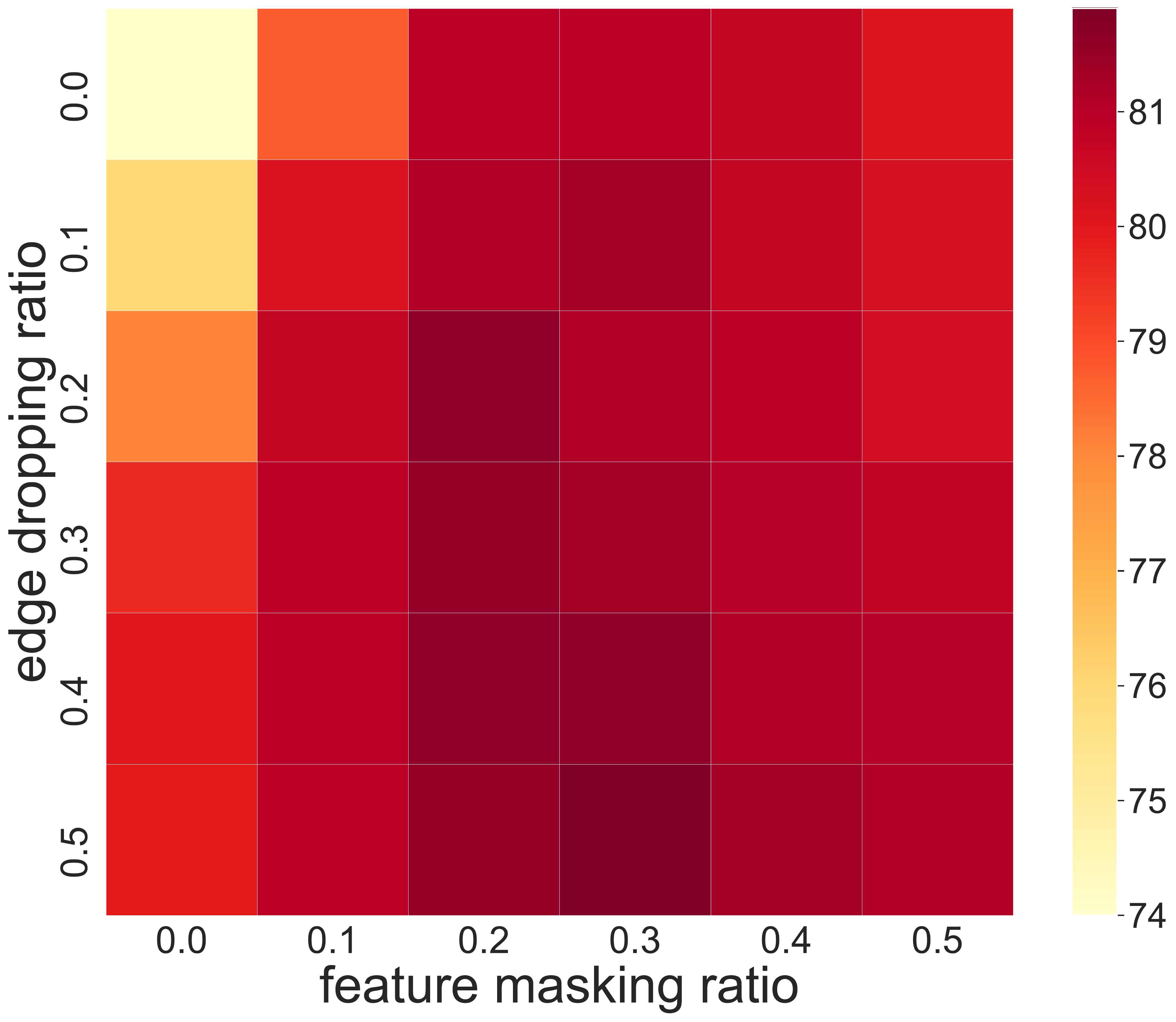}
\end{minipage}
}

\subfigure[Amazon-Computer]{
\begin{minipage}[t]{0.23\linewidth}
\centering
\includegraphics[width=1.0\textwidth,angle=0]{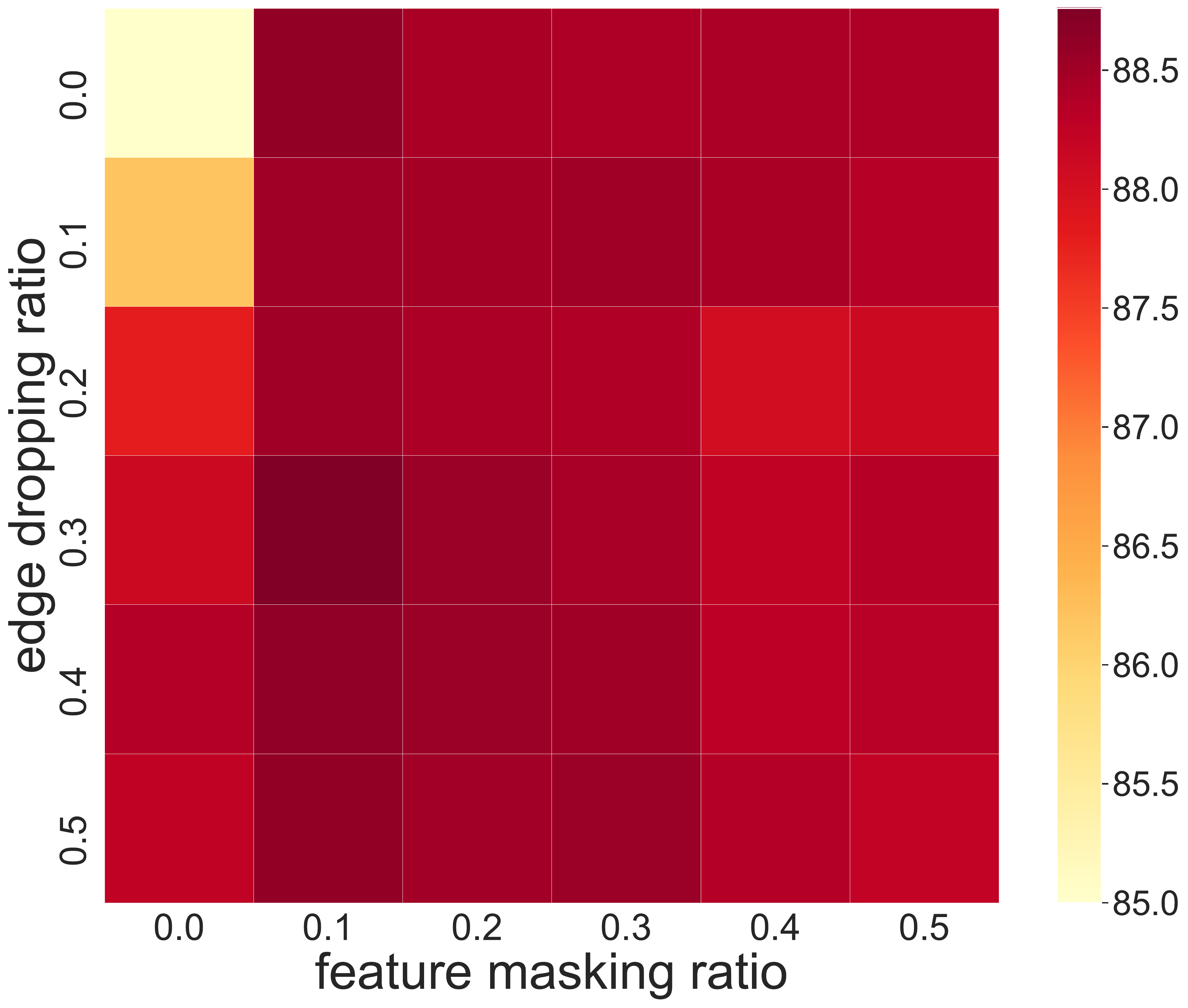}
\end{minipage}
}
\subfigure[Amazon-Photo]{
\begin{minipage}[t]{0.23\linewidth}
\centering
\includegraphics[width=1.0\textwidth,angle=0]{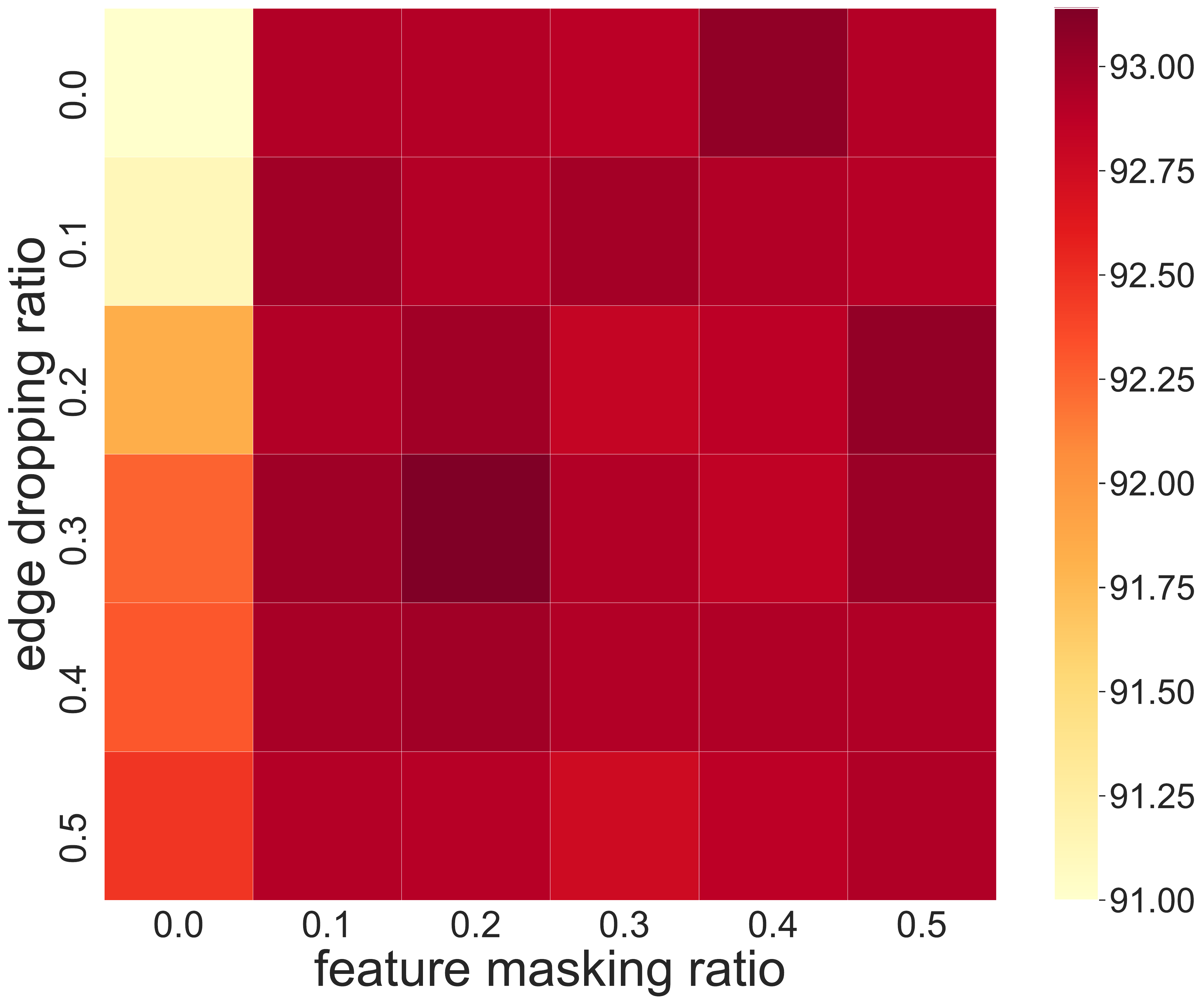}
\end{minipage}
}
\subfigure[Coauthor-CS]{
\begin{minipage}[t]{0.23\linewidth}
\centering
\includegraphics[width=1.0\textwidth,angle=0]{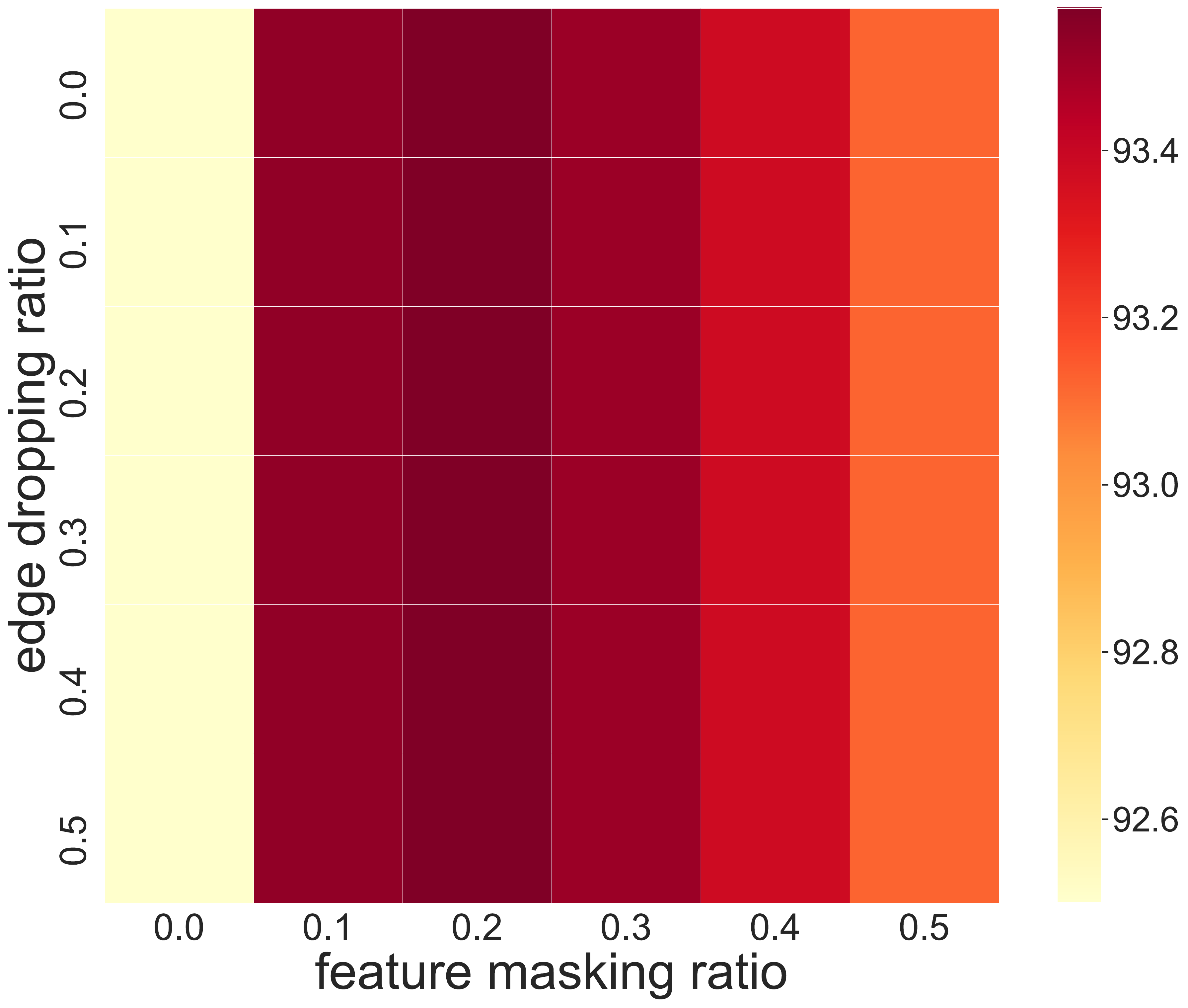}
\end{minipage}
}
\subfigure[Coauthor-Physics]{
\begin{minipage}[t]{0.23\linewidth}
\centering
\includegraphics[width=1.0\textwidth,angle=0]{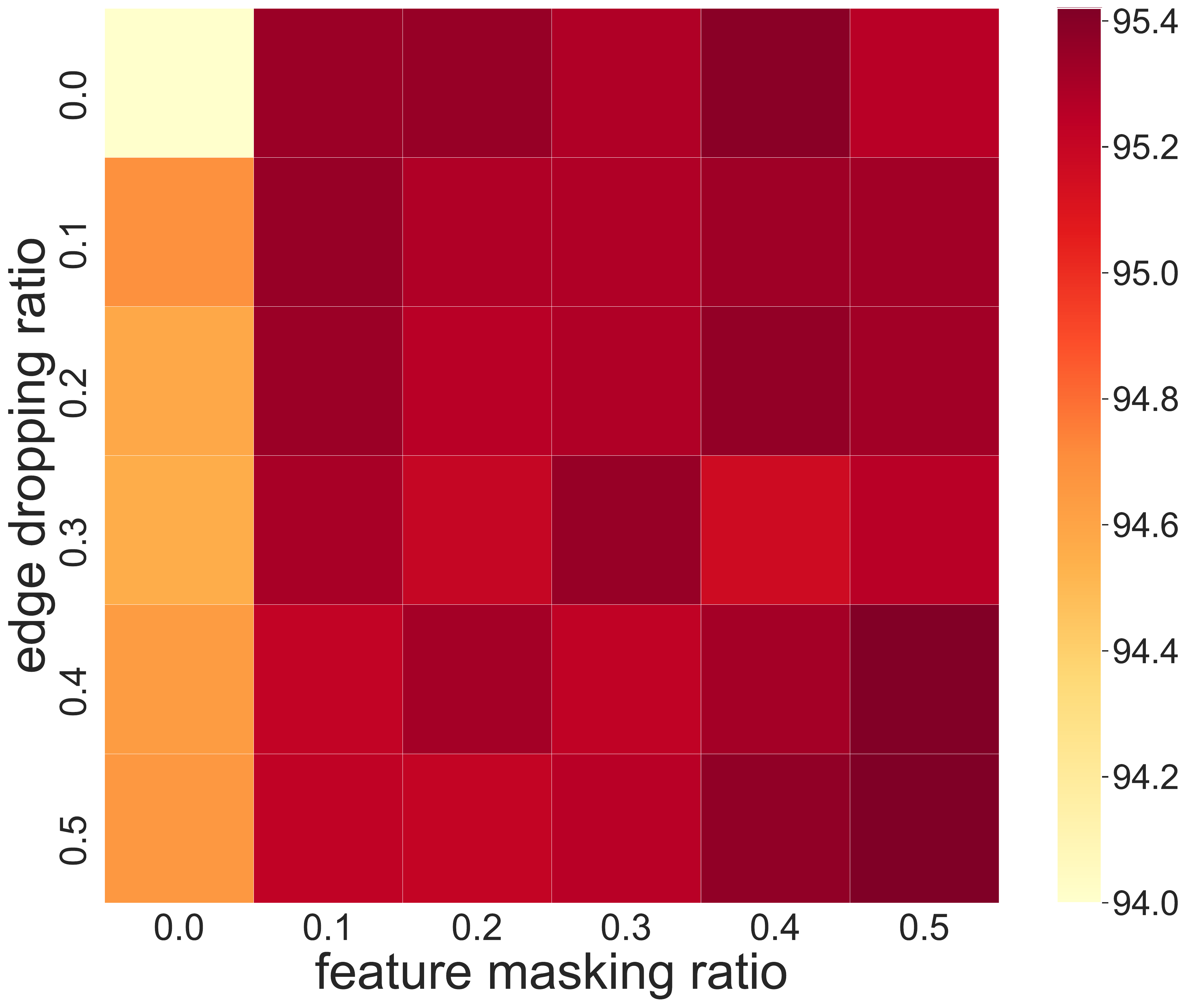}
\end{minipage}
}
\caption{Visualizations of the effects of different augmentation intensity, by adopting different combinations of feature masking ratio $p_f$ and edge dropping ratio $p_e$, and we report test accuracy (\%). Each row represents a specific setting for edge dropping ratio $p_e$ and each column represents a specific setting for feature masking ratio $p_f$. Note that when $p_e = p_f = 0$ (the upper left entry in each subfigure), the test accuracy for different datasets should be: 50.2 on \textit{Cora}; 30.5 on \textit{Citeseer}; 46.4 on \textit{Pubmed}; 54.56 on \textit{Computer}; 83.95 on \textit{Photo}; 90.4 on \textit{CS}; 87.85 on \textit{Physics}. Since these results are much worse than the others, we raise their values in each subfigure for better visualization. On \textit{CS}, the edge dropping ratio $p_e$ would make no difference to the performance as we use MLP as the encoder, which does not take graph structure as input.} 
\label{fig:vis-drop_ratio}
\end{figure}

We further explore the effects of augmentation intensity on downstream node classification tasks. We try different combinations of the feature masking ratio $p_f$ and edge dropping ratio $p_e$, and report the corresponding performance on the 7 benchmarks mentioned in Appendix~\ref{append:exp-dataset}. Other hyper-parameters are the same as reported in Table~\ref{tbl-para}. As we can see in Fig.~\ref{fig:vis-drop_ratio}, for each dataset there exists an optimal $p_e$/$p_f$ combination, that could help the model reach the best performance. Also, we find that our method is not that sensitive to the augmentation intensity: as long as $p_e$ and $p_f$ are in a proper range, our method could still achieve impressive and competitive performance. However, it is still very important to select a proper augmentation intensity as well as augmentation method, in order for label-invariant data augmentations for learning informative representations.

\subsection{Performance under Low Label Rates}
We further evaluate the node embeddings learned through CCA-SSG on downstream node classification tasks (still using linear, logistic regression), with respect to various label rates (ratio of training nodes). The experiments are conducted on three citation networks: \textit{Cora, Citeseer} and \textit{Pubmed}. In the linear evaluation step, we follow the setups in~\cite{deeper-insight}: we train the linear classifier with 1\%, 2\%, 3\%, 4\%, 5\% (resp. 0.05\%, 0.1\%, 0.3\%) training nodes on \textit{Cora} and \textit{Citeseer} (resp. \textit{Pubmed}), and then test the model with another 1000 nodes. Both training nodes and testing nodes are randomly selected for each trial, and we report the mean accuracy through $20$ trials with random splits and random initialization in Table~\ref{tbl-exp-rate}.

We compare our method with Label Propagation, GCN with Chebyshev filter(Cheby) and the vanilla GCN~\cite{gcn}, whose results are taken from~\cite{deeper-insight} as well. As we can see in Table~\ref{tbl-exp-rate}, our method achieves very impressive performance under low label rates, especially when the labeled nodes are really scarce (i.e. 1\% on \textit{Cora} and \textit{Citeseer}, 0.05\% on \textit{Pubmed}). This is because through self-supervised pretraining, our method could fully utilize the information of unlabeled nodes, and learn good representations for them, which make them easy to distinguish even with only a few number of labeled nodes for training.

\begin{table}[t]
	\centering
	\caption{Node classification Accuracy under low label rates (\%).}
	\label{tbl-exp-rate}
	\footnotesize 
	\begin{threeparttable}
{ 
    \setlength{\tabcolsep}{1.6mm}
    {
		\begin{tabular}{c|ccccc|ccccc|ccc}
		    \toprule[1pt]
		 	 Dataset & \multicolumn{5}{c|}{Cora} &  \multicolumn{5}{c|}{Citeseer} & \multicolumn{3}{c}{Pubmed} \\
			 \midrule[0.5pt]
			  Label Rate & 1\% & 2\% & 3\% & 4\% & 5\% & 1\% & 2\% & 3\% & 4\% & 5\% & 0.05\% & 0.1\% & 0.3\% \\
			 \midrule[0.5pt]
			  LP & 62.3 & 65.4 & 67.5 & 69.0 & 70.2 & 40.2 & 43.6 & 45.3 & 46.4 & 47.3 & 66.4 & 65.4 & 66.8  \\
			  Cheby & 52.0 & 62.4 & 70.8 & 74.1 & 77.6 & 42.8 & 59.9 & 66.2 & 68.3 & 69.3  & 47.3  & 51.2 & 72.8  \\
			  GCN & 62.3 & 72.2 & 76.5 & 78.4 & 79.7 & 55.3 & 64.9 & 67.5 & 68.7 & 69.6  & 57.5  & 65.9 & 77.8 \\
    		 \midrule[0.5pt]
    		  CCA-SSG & \textbf{72.5} & \textbf{79.3} & \textbf{81.0} & \textbf{82.0}  & \textbf{82.3} & \textbf{58.9} & \textbf{65.6} & \textbf{68.6} & \textbf{70.8} & \textbf{71.7} & \textbf{68.8} & \textbf{73.1} & \textbf{81.1} \\
			\bottomrule[1pt]
		\end{tabular}}
		}
	\end{threeparttable}
\end{table}

\section{Further Comparisons with previous contrastive methods}\label{append:compare}
In Table~\ref{tbl-comparison} we have made a thorough comparison with typical contrastive methods from the \textbf{technical details}. Here, we further compare our method with more existing contrastive self-supervised graph models (both node-level and graph-level) from the perspective of their general, conceptual designs: 1) How they generate views. 2) The pairs for contrasting. 3) The loss function. 4) Downstream tasks (i.e., node-level, edge-level or graph-level). The comparison is shown in Table~\ref{tbl-appendix-comparison}. Note that this is a high-level comparison with general taxonomy, and each method may have distinct implementation details and specific designs.
\begin{table}[tb!]
	\centering
	\caption{Further conceptual comparison with existing contrastive learning methods on graphs. \textit{View generation in general} (how the method generate views): \underline{Cross-scale} means this method treat elements in different scales of the graph data as different views (e.g. node and graph); \underline{Fix-Diff} means using fixed graph diffusion~\cite{diffusion} operation to create another view; \underline{Rand-Aug} means using random graph augmentations (e.g. edge dropping, feature masking, etc.) to generate views. \textit{Pairs} represents the contrasting components, where $N$ is node and $G$ is graph. \textit{Loss} (i.e. the used loss function): \underline{NCE} represents Noise-Contrastive Estimation~\cite{nce}; \underline{JSD} represents Jensen-Shannon Mutual Information Estimator~\cite{f-gan}; \underline{InfoNCE} represents InfoNCE Estimator~\cite{cpc};  \underline{MINE} means Mutual Information Neural Estimator~\cite{mine}; \underline{BYOL} means the asymmetric objective proposed in the BYOL paper~\cite{byol}. \textit{Tasks} denotes the downstream tasks (node-level, graph-level or edge-level) to which the method has been applied.}
	\label{tbl-appendix-comparison}
	\small
	\begin{threeparttable}
    {{
    		\begin{tabular}{c|lccccc}
			\toprule[0.8pt]
			& Methods & View generation in general & Pairs &  Loss & Tasks   \\
			\midrule[0.6pt]
           \multirow{12}{*}{\rotatebox{90}{Instance-level}} 
            & DGI~\cite{dgi} & Cross-scale  & N-G & NCE & Node              \\
            & InfoGraph~\cite{infograph} & Cross-scale  & N-G  & JSD  & Graph               \\
            & MVGRL~\cite{mvgrl} & Fix-Diff + Cross-scale & N-G & NCE/JSD   & Node/Graph      \\
            & GCC~\cite{gcc}    & Rand-Aug & N-N  & InfoNCE & Node/Graph           \\
            & GMI~\cite{gmi}  & Hybrid$^{1}$  & Hybrid & MINE/JSD & Node/Edge         \\
            & GRACE~\cite{grace} & Rand-Aug & N-N & InfoNCE & Node             \\
            & GraphCL~\cite{graphcl} & Rand-Aug & G-G & InfoNCE & Graph         \\
            & GCA~\cite{grace-ad}  & Rand-Aug & N-N & InfoNCE & Node           \\
            & CSSL~\cite{cssl}  & Rand-Aug & G-G & InfoNCE & Graph           \\
            & IGSD~\cite{igsd} & Rand-Aug & G-G & BYOL+InfoNCE & Graph       \\
            & GraphLog~\cite{graph-log} & Rand-Aug & N-P-G$^2$ & InfoNCE & Graph       \\
            & BGRL~\cite{bgrl}  & Rand-Aug & N-N & BYOL & Node           \\
            & MERIT~\cite{merit}  & Fix-Diff + Rand-Aug & N-N & BYOL+InfoNCE & Node           \\
            \midrule[0.5pt]
            & CCA-SSG (Ours)   & Rand-Aug & F-F & CCA  & Node  \\
			\bottomrule[0.8pt]
		\end{tabular}
		\begin{tablenotes}
		\item[1] The view generation and contrasting pairs in GMI~\cite{gmi} is unique and complex, and could not be classified into any category. 
		\item[2] P denotes hierarchical prototype and could be seen as clustering centroid.
        \end{tablenotes}
		}
		}
	\end{threeparttable}
\end{table}

We highlight that all of the previous methods focus on contrastive learning at instance level. Our paper proposes a non-contrastive and non-discriminative objective as a new self-supervised representation learning framework, inspired by canonical correlation analysis.
\end{document}